\documentclass[pdflatex,sn-nature,Numbered]{sn-jnl}
\usepackage[T1]{fontenc}
\usepackage{geometry}
\usepackage{setspace}
\usepackage{amsmath,amssymb,amsfonts,bm}
\usepackage{bbm}
\usepackage{booktabs,array,multirow,makecell,longtable}
\usepackage[table]{xcolor}
\usepackage{graphicx}
\usepackage{subcaption}
\usepackage{caption}
\usepackage{enumitem}
\usepackage{float}
\usepackage{placeins}
\usepackage{xspace}
\usepackage{siunitx}
\usepackage{threeparttable}
\usepackage{adjustbox}
\usepackage{ragged2e}
\usepackage[most]{tcolorbox}
\usepackage{pifont}

\hypersetup{citecolor=black,linkcolor=black,urlcolor=black}
\makeatletter
\def\email#1{\global\advance\emailcnt by 1\relax%
\if@corauemail%
   \g@addto@macro\corrauthemail{\setcounter{footnote}{0}#1;\ }%
\else%
   \g@addto@macro\authemail{\setcounter{footnote}{0}#1;\ }%
\fi}
\makeatother
\graphicspath{{figures/}{imgs/}{./figures/}{./imgs/}}

\newcommand{\ProjectName}{SCALE\xspace}
\newcommand{\SZ}[1]{}

\title[SCALE]{\texorpdfstring{\ProjectName: Scalable Conditional Atlas-Level Endpoint Transport for virtual cell perturbation prediction}{SCALE: Scalable Conditional Atlas-Level Endpoint Transport for virtual cell perturbation prediction}}

\author[1,2]{\fnm{Shuizhou} \sur{Chen}}\equalcont{These authors contributed equally to this work.}
\author[1]{\fnm{Lang} \sur{Yu}}\equalcont{These authors contributed equally to this work.}
\author[3,4,5]{\fnm{Xueqin} \sur{Lin}}\equalcont{These authors contributed equally to this work.}
\author[1,6,7]{\fnm{Xinjie} \sur{Mao}}\equalcont{These authors contributed equally to this work.}
\author[8]{\fnm{Songming} \sur{Zhang}}\equalcont{These authors contributed equally to this work.}
\author[1]{\fnm{Xinyu} \sur{Gu}}\equalcont{These authors contributed equally to this work.}

\author[1,9]{\fnm{Hao} \sur{Wu}}
\author[1,9]{\fnm{Sheng} \sur{Xu}}
\author[1,10]{\fnm{Kedu} \sur{Jin}}
\author[1]{\fnm{Lei} \sur{Bai}}

\author*[2]{\fnm{Quan} \sur{Qian}}\email{qqian\@shu.edu.cn}
\author*[11]{\fnm{Qin} \sur{Chen}}
\author*[3,4,5]{\fnm{Qiang} \sur{Gao}}
\author*[1,9]{\fnm{Siqi} \sur{Sun}}
\author*[1]{\fnm{Zhangyang} \sur{Gao}}\email{gaozhangyang\@pjlab.org.cn}

\affil[1]{\orgname{Shanghai Artificial Intelligence Laboratory}, \orgaddress{\city{Shanghai}, \country{China}}}

\affil[2]{\orgdiv{School of Computer Engineering \& Science}, \orgname{Shanghai University}, \orgaddress{\city{Shanghai}, \country{China}}}

\affil[3]{\orgdiv{Department of Hepatobiliary Surgery \& Transplantation, Liver Cancer Institute and State Key Laboratory of Genetics and Development of Complex Phenotypes}, \orgname{Zhongshan Hospital, Fudan University}, 
\orgaddress{\city{Shanghai}, \country{China}}}

\affil[4]{\orgdiv{Key Laboratory of Carcinogenesis \& Cancer Invasion, Ministry of Education; Institutes of Biomedical Sciences}, \orgname{Human Phenome Institute, Fudan University}, 
\orgaddress{\city{Shanghai}, \country{China}}}

\affil[5]{\orgdiv{Shanghai Academy of Natural Sciences}, \orgaddress{\city{Shanghai}, \country{China}}}

\affil[6]{\orgdiv{Shanghai Innovation Institute}, \orgaddress{\city{Shanghai}, \country{China}}}

\affil[7]{\orgdiv{Westlake University}, \orgaddress{\city{Hangzhou}, \country{China}}}

\affil[8]{\orgdiv{Faculty of Computer Science and Control Engineering}, \orgname{Shenzhen University of Advanced Technology}, \orgaddress{\city{Shenzhen}, \country{China}}}

\affil[9]{\orgdiv{Research Institute of Intelligent Complex Systems}, \orgname{Fudan University}, \orgaddress{\city{Shanghai}, \country{China}}}

\affil[10]{\orgdiv{School of Life Science and Technology}, \orgname{China Pharmaceutical University}, \orgaddress{\city{Nanjing}, \country{China}}}

\affil[11]{\orgdiv{School of Computer Science and Technology}, \orgname{East China Normal University}, \orgaddress{\city{Shanghai}, \country{China}}}

\abstract{%

    Virtual-cell models aim to predict how cell populations respond to perturbations, but control and treated cells are measured as unpaired populations, complicating the learning of perturbation-specific effects. We present SCALE, a conditional transport model that represents cells as unordered sets and predicts treated populations without cell-level matching. A shared set-aware encoder and conditional DiT backbone learn latent transport, making endpoint supervision directly delta-aligned without an auxiliary delta objective. Across genetic, chemical, developmental and immune perturbations, SCALE recovered gene-expression changes, response directions and population structure. In CRISPR data with dominant cell-line effects, SCALE outperformed competing methods across seven metrics and maintained separation among gene-target representations rather than collapsing them into a shared region. SCALE further prioritized cytokines predicted to produce distinct immune activation and inflammatory responses. Experiments using matched PBMC samples from three donors confirmed these predicted differences. Together, SCALE enables perturbation-specific prediction from unpaired populations and supports experimental prioritization.

}

\keywords{virtual cell, perturbation prediction, conditional transport, single-cell foundation model, BioNeMo, Tahoe-100M}

\begin{document}
\maketitle

\section{Introduction}
\label{sec:introd}

Virtual-cell models aim to predict how cells respond to genes, drugs or cytokines before an experiment is performed\cite{Bunne2024VirtualCell,Rood2024PerturbationAtlas}. Most single-cell assays destroy the cells they measure. An experiment therefore observes one group of untreated cells and a different group of treated cells, rather than the same cells before and after treatment. The prediction target is thus a population: given control cells and a perturbation, predict the treated cells. A useful prediction should capture the shift in average expression, the range of responses and the presence of distinct response states. These population features are the main information that single-cell data provide beyond a single average profile.

Many methods treat the perturbation task as a cell-to-cell prediction problem. The model takes one control cell and predicts one treated cell\cite{lotfollahi2019scgen,lotfollahi2023cpa,roohani2024gears}. However, single-cell experiments measure different cells before and after treatment, so the true pairs remain unknown. Researchers must create these pairs either at random or with a matching method. As each pairing method can create a different target, the model may learn the pairing rule instead of the difference between the two populations. Cell-level training may also cause predicted cells to converge toward the average response. This may result in good mean-expression scores even though the variation among treated cells is lost. The predicted change from one cell to another is not a measured biological path, because the experiment does not track the same cell through treatment. A more direct task is therefore to predict the treated population from the control population without assuming cell pairs.

\noindent\textbf{The challenge is to combine a correct population-level target with strong perturbation-specific supervision.}
Population-level distribution learning matches the structure of unpaired single-cell experiments, but this formulation does not guarantee perturbation-specific predictions. Systema showed that complex predictors can recover broad control-to-treated differences while missing perturbation-specific effects\cite{vinas2026systema}. Diversity by Design and Departures linked this failure to mode collapse, sparse response signals, losses that dilute these signals and metrics that favor the dataset mean\cite{mejia2025diversity, chi2026departures}. By contrast, cell-to-cell models assume pairs that the experiments do not provide, yet they can achieve strong predictive performance\cite{lotfollahi2019scgen,lotfollahi2023cpa,roohani2024gears}. One possible explanation is that pointwise losses provide a more direct training signal than distribution-level losses, even when the cell-level targets are artificial. Strong cell-to-cell performance therefore does not validate the pairing assumption; instead, it exposes weak perturbation-specific supervision in distribution-based learning. The central problem is to learn the correct population endpoint without sacrificing perturbation-specific information.

\noindent\textbf{Evaluation must test perturbation specificity, not only mean-expression accuracy.}
Recent benchmarks show that complex models do not always beat an average treatment response or a simple additive baseline under many contexts \cite{ahlmann2025linear,vinas2026systema,wei2026benchmarking}. These findings leave two separate questions. Is exact cell correspondence necessary for learning a useful endpoint map? Can an unpaired population model preserve perturbation-specific response instead of compressing distinct perturbations towards a shared mean? Low error in mean expression is not enough. A useful model should distinguish perturbations, identify the main response genes and preserve whether their expression increases or decreases. It should also retain part of the range and internal differences of the treated population. Most importantly, it should help select experiments expected to produce different biological outcomes.

\noindent\textbf{SCALE makes population-level prediction tractable through a set-aware latent representation.}
Predicting this population change directly from thousands of genes is difficult. Single-cell data contain many zeros, technical noise and genes that are weakly related to the tested perturbation. One measured population can also contain several cell states. SCALE first converts each cell into a lower-dimensional representation, called a latent space. It then predicts how the control population changes under the perturbation to match the treated endpoint. This representation keeps the main cell states and population structure while making population-level prediction easier. The path used during training is only a computational tool. It does not describe the biological events between the control and treated measurements.

\noindent\textbf{SCALE predicts unpaired endpoints without losing perturbation specificity.}
Here we introduce SCALE, a model that predicts a treated cell population from separately measured control and treated endpoints. It trains directly on these populations and does not assume one-to-one cell pairs. We test SCALE on gene perturbations, drugs, cytokines and developmental data. BioNeMo supports training and prediction on large datasets\cite{stjohn2024bionemo}. Methods such as scVI, scGPT, CellOT, STATE and PerturbDiff provide important foundations\cite{lopez2018scvi,cui2024scgpt,bunne2023cellot,adduri2025state,yuan2026perturbdiff}. Our central test is whether direct population-level training can meet both requirements: no unobserved cell pairing and no collapse of perturbation-specific information into an average response.

\noindent\textbf{SCALE recovers perturbation-specific responses and uses them to guide cytokine experiments.}
SCALE beat average-response baselines and existing models in several tasks, while no method was best on every dataset and metric. In CRISPR screens, SCALE remained target-specific responses after cell-line and treatment averages were removed, and its latent responses retained distinct directions and spread. In drug and developmental data, it better preserved changes in treated populations and showed where predictions remained weak. In immune data, SCALE predicted cytokine responses across donors, cell types and patients excluded from training on different datasets. We used one set of the out-of-context predictions to select cytokines expected to elicit distinct immune effects and then tested those choices in donor-matched PBMC experiments. Those experiments revealed distinct patterns of CD8 T-cell and NK-cell activation, together with differences in the secretion of inflammatory molecules. This prediction-to-validation sequence therefore tests whether population-level models can move beyond computational performance to guide experiments toward biologically distinct outcomes.


\section{Results}

\subsection{SCALE recovers perturbation-specific responses accurately by learning transport between unpaired single-cell endpoints}
\label{sec:latent_population_transfer}

\begin{figure}[H]
    \centering
    \includegraphics[width=1\textwidth]{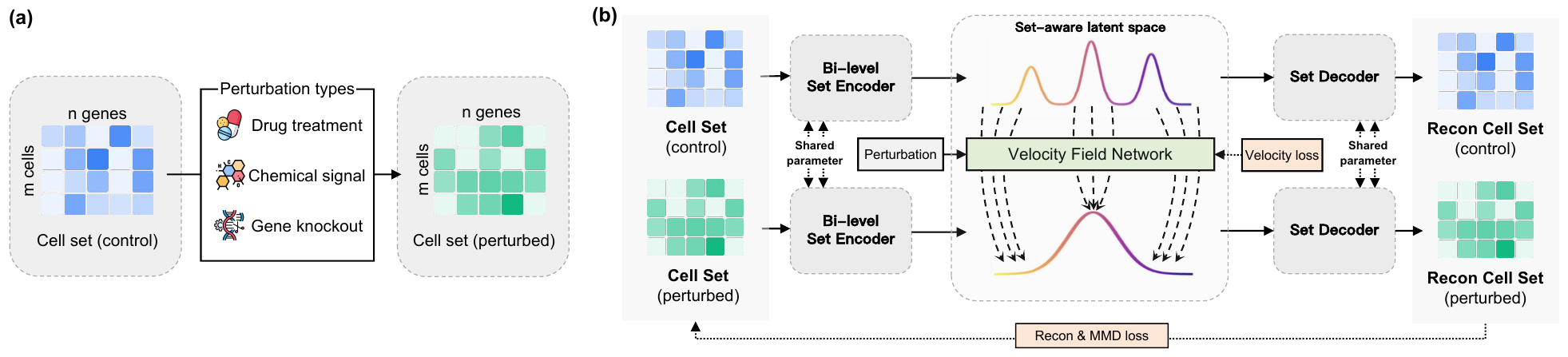}
    \caption{\textbf{Overview of \ProjectName\ (a, prediction task; b, model design).} \textbf{(a) Population-level prediction.} Given separate control and treated populations from the same experiment, the model predicts the treated population from the control population and perturbation. It does not assume that control and treated cells are paired. \textbf{(b) SCALE architecture.} SCALE represents each unordered population in a lower-dimensional, set-aware space and predicts the treated endpoint. The training system supports large datasets, and the evaluation tests whether the prediction is specific to each perturbation.}
    \label{fig:my_image}
\end{figure}

Single-cell perturbation experiments measure separate groups of untreated and treated cells. They do not track the same cell before and after treatment. SCALE therefore takes an unordered group of control cells, the perturbation and the measured context as input. It encodes the cells in an order-independent, lower-dimensional representation, called a set-aware latent space, and predicts the treated group (Figure~\ref{fig:my_image}). Training compares this prediction with the measured treated cells and does not require one-to-one cell matching.

This learned transport path is only a way to link the untreated and treated endpoint. It is not a measured biological trajectory because the data contain neither intermediate time points nor before--after tracks for individual cells. We therefore asked whether SCALE could reproduce the centre, direction and internal variation of each treated population instead of fitting the same average response for many perturbations.

We evaluated SCALE in three well-known benchmarks. Where required by the benchmark, all models predicted the same 2,000 highly variable genes~\cite{adduri2025state,yuan2026perturbdiff}:

\begin{itemize}
    \item \textbf{PBMC} contains 90 cytokine conditions across 12 donors and 16 major cell types. It tests immune-response prediction across donors and cell types~\cite{oesinghaus2025cytokine}.

    \item \textbf{Tahoe-100M} contains more than 100 million profiles from 50 cancer cell lines and over 1,100 treatment conditions. It tests drug-response prediction at large scale~\cite{zhang2025tahoe100m}.

    \item \textbf{Replogle-Nadig} contains 2,024 filtered CRISPRi perturbations across four human cell lines. It tests whether the model can detect small gene-specific responses despite large differences between cell lines~\cite{replogle2022mapping,nadig2025trade}.
\end{itemize}

Each dataset tests a different kind of generalization, although all follow their published normalization and gene-selection procedures. Methods provides the full details. In PBMC, training includes 30\% of the cytokine conditions from four evaluation donors and testing uses the remaining 70\%. This split predicts missing cytokine conditions for partly measured donors, not completely new donors. Tahoe leaves five biologically distinct cell lines out of training. Replogle tests one cell line after training on only some perturbations from that line. These splits answer different questions and should not be treated as one generic test of new conditions.

BioNeMo made SCALE faster both per training step and per processed cell. We stored sparse expression matrices in LMDB files grouped by experimental condition, retrieved matched controls separately and ran training in BioNeMo\cite{stjohn2024bionemo} with sampling that accounts for each batch. Using each framework's own definition of an epoch, SCALE reduced epoch time from 1298.43~s to 243.00~s. It also increased iteration rate from 0.1540 to 1.9259 iterations per second, a 12.51$\times$ gain. Because the two sampling schemes process different numbers of cells per epoch, cells per second provide the fairest workload comparison. SCALE processed 27,818.20 cells/s, compared with 21,584.40 cells/s for STATE, a 1.29$\times$ gain (Table~\ref{tab:system_acceleration}). These numbers compare the complete systems and do not measure the data loader alone.

{\centering
\small
\captionof{table}{\textbf{Training and prediction speed of the BioNeMo-based \ProjectName system.}
Training speed uses each framework's own definition of an epoch. Prediction speed is normalized by the number of cells processed and is compared with STATE.}
\label{tab:system_acceleration}
\setlength{\tabcolsep}{5pt}

\resizebox{\textwidth}{!}{
\begin{tabular}{llcccccc}
\toprule
& & \multicolumn{3}{c}{\textbf{Pretraining}} & \multicolumn{3}{c}{\textbf{Inference}} \\
\cmidrule(lr){3-5} \cmidrule(lr){6-8}
\multicolumn{1}{c}{\textbf{Model}}
& \multicolumn{1}{c}{\textbf{Infra}}
& \makecell{\textbf{Native epoch}\\\textbf{time (s) $\downarrow$}}
& \textbf{Iter/s $\uparrow$}
& \makecell{\textbf{Speedup vs.\ \textsc{STATE}}\\\textbf{(iter/s) $\uparrow$}}
& \textbf{Cells/epoch $\uparrow$}
& \makecell{\textbf{Cell-normalized }\\\textbf{throughput (cells/s) $\uparrow$}}
& \makecell{\textbf{Speedup vs.\ \textsc{STATE}}\\\textbf{(throughput) $\uparrow$}} \\
\midrule
\textsc{STATE}-280M
& Original
& 1298.43
& 0.1540
& 1.00$\times$
& 4,705,402
& 21,584.40
& 1.00$\times$ \\
{\ProjectName}-184M
& BioNeMo
& \textbf{243.00}
& \textbf{1.9259}
& \textbf{12.51}$\times$
& 5,187,584
& \textbf{27,818.20}
& \textbf{1.29}$\times$ \\
\bottomrule
\end{tabular}
}
\par}

Across three benchmarks, SCALE recovered more than an average treatment effect, although no method was best on every metric. Table~\ref{tab:table1_unified_benchmark} reports seven metrics. It combines public or locally reproduced CellEval results with VCBench predictions that we re-evaluated using CellEval~0.6.6. The rows do not all use the same prediction pipeline or set of conditions, so the table note states the source and coverage of every result.

\begin{table*}[!ht]
\centering
\scriptsize
\caption{\textbf{Three-dataset, seven-metric perturbation benchmark.}
Bold denotes the best value and underlining the second best within each dataset. MSE and MAE are minimized and the other metrics are maximized.}
\label{tab:table1_unified_benchmark}

\setlength{\tabcolsep}{3.2pt}
\renewcommand{\arraystretch}{1.0}
\resizebox{\textwidth}{!}{
\begin{tabular}{lccccccc}
\toprule
Model & MSE $\downarrow$ & MAE $\downarrow$ & PDCorr $\uparrow$
& DEOver $\uparrow$ & DEPrec $\uparrow$ & LFCSpear $\uparrow$ & DirAgr $\uparrow$ \\
\midrule

\rowcolor[HTML]{EFEFEF}
\multicolumn{8}{c}{\textbf{\textit{Replogle}}} \\
{\ProjectName} (\textit{ours})
& \textbf{0.0043} & \textbf{0.0390} & \textbf{0.7510}
& \textbf{0.4040} & \textbf{0.2540} & \textbf{0.7960} & \textbf{0.9220} \\
STATE
& \underline{0.0071} & \underline{0.0573} & \underline{0.4132}
& \underline{0.2294} & \underline{0.2128} & \underline{0.4682} & \underline{0.7585} \\
Linear Additive & 0.1798 & 0.3901 & 0.1530 & 0.0440 & 0.0907 & 0.1254 & 0.5197 \\
Latent Additive & 0.3286 & 0.4104 & 0.1095 & 0.1417 & 0.0992 & 0.0968 & 0.5658 \\
Decoder Only & 0.2842 & 0.4864 & 0.0974 & 0.0492 & 0.0977 & 0.1057 & 0.5113 \\
CPA & 0.0130 & 0.0723 & 0.1669 & 0.0817 & 0.0774 & 0.1766 & 0.5983 \\
PRnet & 0.0114 & 0.0679 & 0.2644 & 0.1067 & 0.0923 & 0.3039 & 0.6510 \\
BioLORD & 0.0130 & 0.0724 & 0.1672 & 0.0812 & 0.0783 & 0.1800 & 0.5987 \\
GenePert & 0.0827 & 0.1904 & 0.0589 & 0.0923 & 0.0985 & 0.0881 & 0.5471 \\
SAMS-VAE & 0.0777 & 0.1817 & 0.0717 & 0.1129 & 0.0964 & 0.1198 & 0.5591 \\
scLAMBDA & 0.0217 & 0.0926 & 0.0911 & 0.0959 & 0.0905 & 0.1363 & 0.5686 \\
GEARS & 0.0115 & 0.0709 & 0.2311 & 0.0841 & 0.0817 & 0.2537 & 0.6405 \\
\midrule

\rowcolor[HTML]{EFEFEF}
\multicolumn{8}{c}{\textbf{\textit{PBMC}}} \\
{\ProjectName} (\textit{ours})
& \underline{0.0067} & \underline{0.0320} & \textbf{0.6430}
& \textbf{0.4430} & \textbf{0.5280} & 0.2270 & 0.6790 \\
STATE
& \textbf{0.0024} & \textbf{0.0190} & 0.4200
& 0.1100 & 0.1300 & 0.1700 & 0.5800 \\
Linear Additive & 0.1596 & 0.1771 & 0.0012 & \underline{0.2104} & 0.2250 & 0.2190 & 0.4872 \\
Latent Additive & 0.1555 & 0.2292 & 0.0258 & 0.0742 & 0.2168 & 0.3575 & 0.6330 \\
Decoder Only & 0.2730 & 0.2447 & 0.0932 & 0.1736 & 0.2168 & 0.3343 & 0.6456 \\
CPA & 0.0094 & 0.0376 & \underline{0.4820} & 0.1540 & 0.2285
& \underline{0.3656} & \textbf{0.7153} \\
PRnet & 0.0170 & 0.0596 & 0.1494 & 0.1364 & \underline{0.2484} & -0.1140 & 0.4654 \\
BioLORD & 0.0095 & 0.0382 & 0.4283 & 0.1515 & 0.2298 & 0.3146 & \underline{0.6835} \\
GenePert & 0.2656 & 0.2407 & 0.0796 & 0.1578 & 0.2418 & 0.1182 & 0.5490 \\
SAMS-VAE & 0.2771 & 0.2381 & 0.0313 & 0.1519 & 0.2167 & 0.2862 & 0.6065 \\
scLAMBDA & 0.0089 & 0.0479 & 0.2675 & 0.0879 & 0.2163 & \textbf{0.3745} & 0.6794 \\
GEARS & 0.0139 & 0.0583 & 0.0249 & 0.1219 & 0.2302 & 0.2226 & 0.6047 \\
\midrule

\rowcolor[HTML]{EFEFEF}
\multicolumn{8}{c}{\textbf{\textit{Tahoe-100M}}} \\
{\ProjectName} (\textit{ours})
& \textbf{0.0003} & \textbf{0.0059} & \textbf{0.8584}
& 0.1204 & 0.2480 & 0.3816 & \textbf{0.8092} \\
STATE
& \underline{0.0004} & \underline{0.0067} & \underline{0.8463}
& \textbf{0.3025} & \textbf{0.3937} & 0.3952 & \underline{0.7940} \\
Linear Additive & 0.0071 & 0.0344 & 0.2398 & 0.1289 & 0.2288 & \textbf{0.4350} & 0.6967 \\
Latent Additive & 0.4065 & 0.6126 & 0.1466 & 0.0613 & 0.2293 & 0.3799 & 0.5369 \\
Decoder Only & 0.0124 & 0.0415 & 0.2888 & 0.2236 & 0.2290 & \underline{0.4216} & 0.7219 \\
CPA & 0.0010 & 0.0101 & 0.2706 & 0.1218 & 0.2183 & 0.2019 & 0.6015 \\
PRnet & 0.0021 & 0.0183 & 0.1885 & \underline{0.2902} & \underline{0.3556} & -0.0915 & 0.4660 \\
BioLORD & 0.0011 & 0.0108 & 0.1338 & 0.1074 & 0.2202 & 0.1601 & 0.5513 \\
GenePert & 0.0111 & 0.0397 & 0.2282 & 0.1979 & 0.3209 & -0.0104 & 0.5648 \\
SAMS-VAE & 0.0107 & 0.0390 & 0.2490 & 0.1387 & 0.2290 & 0.4065 & 0.6910 \\
scLAMBDA & 0.0070 & 0.0300 & 0.2685 & 0.1397 & 0.2308 & 0.3592 & 0.6804 \\
GEARS & 0.0011 & 0.0097 & -0.0038 & 0.1021 & 0.2311 & -0.0885 & 0.4797 \\
\bottomrule
\end{tabular}
}
\par\vspace{2pt}
{\raggedright\tiny
The remaining models were pretrained with VCBench and evaluated with CellEval 0.6.6.\par}
\end{table*}
\FloatBarrier

On Replogle, SCALE led all seven metrics. For example, PDCorr was 0.7510 for SCALE and 0.4132 for STATE, while DEOver was 0.4040 and 0.2294, respectively. On PBMC, SCALE had the best PDCorr (0.6430), DEOver (0.4430) and DEPrec (0.5280). STATE had the lowest MSE and MAE, scLAMBDA had the highest LFCSpear, and CPA had the highest DirAgr. On Tahoe-100M, SCALE led MSE (0.0003), MAE (0.0059), PDCorr (0.8584) and DirAgr (0.8092). STATE led DEOver and DEPrec, while Linear Additive led LFCSpear. SCALE was therefore not best on every dataset and metric. Instead, the results show that separately measured control and treated populations contain perturbation-specific information, with each metric capturing a different part of that information. These benchmarks therefore establish the first step of the study: SCALE learns directly from those populations while retaining perturbation-specific signals. The following analyses of CRISPR, drug, developmental and immune experiments test whether those signals correspond to distinct biological effects and can support experimental selection.

\subsection{SCALE recovers target-specific CRISPR responses despite strong cell-line shifts}
\label{sec:replogle_results}

\begin{figure}[p]
\refstepcounter{figure}
\label{fig:replogle}
\addcontentsline{lof}{figure}{\protect\numberline{\thefigure}Replogle CRISPR responses reveal target-specific recovery beyond cell-line averages}
\begingroup
\footnotesize
\setstretch{1.0}
\noindent\textbf{Figure \thefigure\ | SCALE predicts gene-specific CRISPR responses beyond cell-line averages (a--d, task and cell-line structure; e--i, target-specific response recovery; j, response-space coverage across cell lines; k, latent representation spread).}
\par\smallskip
{\centering
\includegraphics[width=0.95\textwidth]{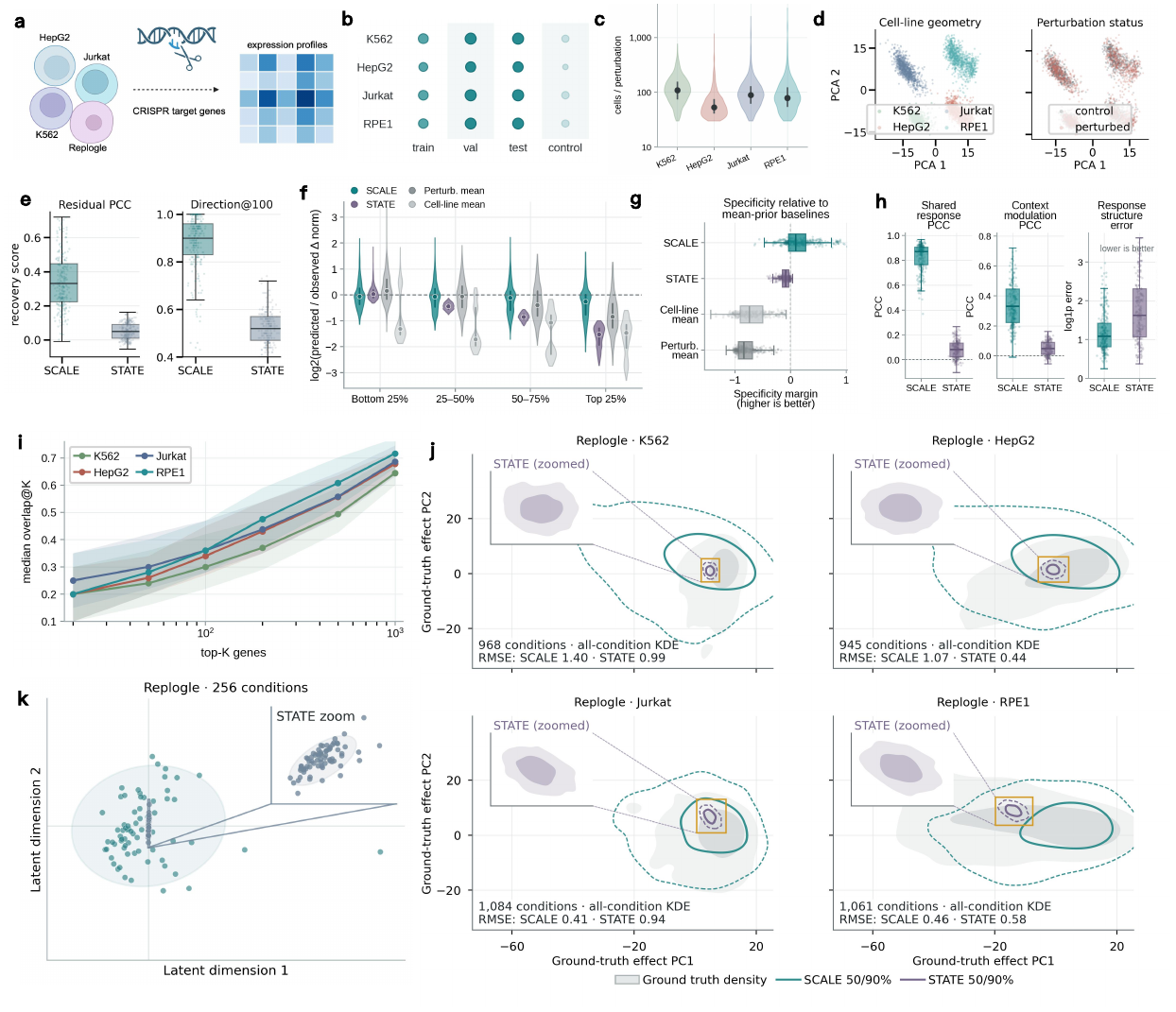}\par}
\par\vspace{1pt}
\scriptsize
\setstretch{0.94}
\justifying
\noindent\textbf{(a) Replogle CRISPR task.} Prediction of target-gene-induced expression profiles across four human cell lines.
\textbf{(b) Data split.} Perturbations and matching controls in the training, validation and test sets.
\textbf{(c) Number of treated cells.} Cell counts per perturbation in each cell line; points show medians.
\textbf{(d) Expression separates mainly by cell line.} Principal-component plots coloured by cell line and perturbation status.
\textbf{(e) Gene-specific response and direction.} Residual PCC and Direction@100 for SCALE and STATE; labels report medians.
\textbf{(f) Predicted response size.} Log$_2$ ratio of predicted to measured response norm across groups of measured perturbation strength; zero marks equal size.
\textbf{(g) Accuracy beyond mean responses.} Margin over the better of the cell-line and perturbation mean predictions; positive values indicate a more target-specific prediction.
\textbf{(h) Shared and cell-line-specific responses.} PCC of the response averaged across cell lines, PCC of each cell line's deviation from that average, and $\log(1+x)$-transformed response-component error; each point is one target measured in at least three cell lines (Appendix~\ref{app:replogle_metrics}).
\textbf{(i) Top-$K$ response-gene overlap.} Median differential-expression overlap across cell lines as the number of selected genes increases.
\textbf{(j) Response-space coverage by cell line.} Ground-truth density and the 50\% and 90\% contours of SCALE- and STATE-predicted responses in PCA space of signed log$_2$ fold-change vectors. Labels report the number of conditions and all-condition KDE RMSE; lower is better.
\textbf{(k) Latent representation spread.} Two-dimensional latent projections for 256 conditions; the compact STATE region is enlarged in the inset.
Each perturbation condition, rather than each cell or gene, is one unit in the overall comparisons. Low-dimensional response and latent projections are descriptive and do not prove that the full treated-cell distribution was recovered.
\par
\endgroup
\end{figure}

The Replogle CRISPR screen\cite{replogle2022mapping,nadig2025trade} measures related gene perturbations in several cell lines. This design tests whether a model predicts the response to the targeted gene or merely copies the cell line's usual expression pattern. We therefore asked whether the models preserved gene-specific changes after removing average effects, which responses were shared across cell lines and where predicted effect sizes remained inaccurate (Figure~\ref{fig:replogle}). This benchmark contained 643,413 cells from K562, HepG2, Jurkat and RPE1. It included 5,763 cell-line-specific perturbation labels and a median of 80 cells per perturbation (Figure~\ref{fig:replogle}a--c).

In this dataset, expression differences among cell lines were much larger than CRISPR effects within each line (Figure~\ref{fig:replogle}d). Large differences among cell lines can push models towards average responses and collapse target-specific representations. This means a model can lower overall error by learning cell-line averages and the mean perturbation effect while missing target-specific changes. Consistent with this failure mode, a 256-condition comparison occupied a broad region in SCALE's latent projection, whereas STATE's representations were confined to the compact cluster enlarged in the inset (Figure~\ref{fig:replogle}k). We therefore tested SCALE in three steps: whether it recovered gene-specific responses beyond both average effects, captured shared and cell-line-specific effects, and kept perturbations distinct in response and latent spaces.

First, SCALE recovered gene-specific CRISPR responses that could not be explained by cell-line or average-treatment effects. To test this claim, we removed cell-line and average-treatment effects and compared the remaining predicted and measured responses. Residual PCC measures how closely the remaining predicted pattern matches the measured pattern. Direction@100 measures whether the 100 leading response genes were predicted to increase or decrease in the correct direction. SCALE achieved higher median values than STATE for both residual PCC (0.33 versus 0.05) and Direction@100 (0.88 versus 0.57; Figure~\ref{fig:replogle}e).

SCALE reproduced response sizes more accurately than STATE and the two mean baselines, especially for strong responses (Figure~\ref{fig:replogle}f). We then compared each model with cell-line and treatment mean baselines to test whether it learned target-specific information beyond either average. SCALE was the only model whose median gene-specific score exceeded both baselines (Figure~\ref{fig:replogle}g). A positive margin means that target-specific predictions were more accurate than both average baselines. Together, the response-size and specificity tests show that SCALE did more than copy an average response.

Second, SCALE captured both shared CRISPR effects and their cell-line-specific changes more accurately than STATE. To separate these two components, we analysed 738 perturbations measured in at least three cell lines, including 194 measured in all four. We first averaged each target's signed log$_2$ fold-change profile across cell lines. The study-specific shared-response PCC correlates the predicted and measured average profiles. Its median was 0.871 for SCALE and 0.082 for STATE (Figure~\ref{fig:replogle}h).

We next subtracted that average from each cell line. The study-specific context-modulation PCC correlates the predicted and measured deviations from the average, and reached a median of 0.332 for SCALE and 0.049 for STATE. We also calculated a response-component error that combines error in shared-response strength with error in the relative strength of cell-line modulation. Lower values are better; the raw median was 1.962 for SCALE and 4.092 for STATE (Figure~\ref{fig:replogle}h). These diagnostics are defined in Appendix~\ref{app:replogle_metrics}. Together, they show that SCALE recovered both the response shared across cell lines and the changes specific to each cell line more accurately than STATE.

Most importantly, SCALE preserved a broader range of condition-specific responses, whereas STATE compressed them into a compact region. In the response projection, SCALE's 50\% and 90\% contours covered more of the measured region across all four cell lines, whereas STATE remained concentrated near a small central region (Figure~\ref{fig:replogle}j). A separate two-dimensional projection of latent embeddings across 256 conditions showed the same pattern. SCALE separated perturbation-specific embeddings across a broad region, whereas STATE embeddings collapsed into the compact area enlarged in the inset (Figure~\ref{fig:replogle}k). Together with the gene-level metrics, these results show that SCALE did more than improve HVG-level accuracy. It preserved target-to-target differences in both its learned representation and predicted responses.

Taken together, the population-level response patterns and gene-level results show that SCALE retained both relationships among perturbations and the genes and directions that defined their biological effects. This resolution enables candidate interventions to be compared by their predicted effects rather than by an average response alone.

\subsection{SCALE preserves drug-response specificity across cell lines, combinations and doses}
\label{sec:sciplex_results}

\begin{figure}[p]
\refstepcounter{figure}
\label{fig:sciplex}
\addcontentsline{lof}{figure}{\protect\numberline{\thefigure}Chemical-response transfer across cell context, combinations and dose}
\begingroup
\footnotesize
\setstretch{1.0}
\noindent\textbf{Figure \thefigure\ | Drug-response prediction across cell lines, drug combinations and doses (a--e, cell-line transfer; f--j, combinations excluded from training; k--q, high-dose prediction).}
\par\smallskip
{\centering
\includegraphics[width=0.9\textwidth]{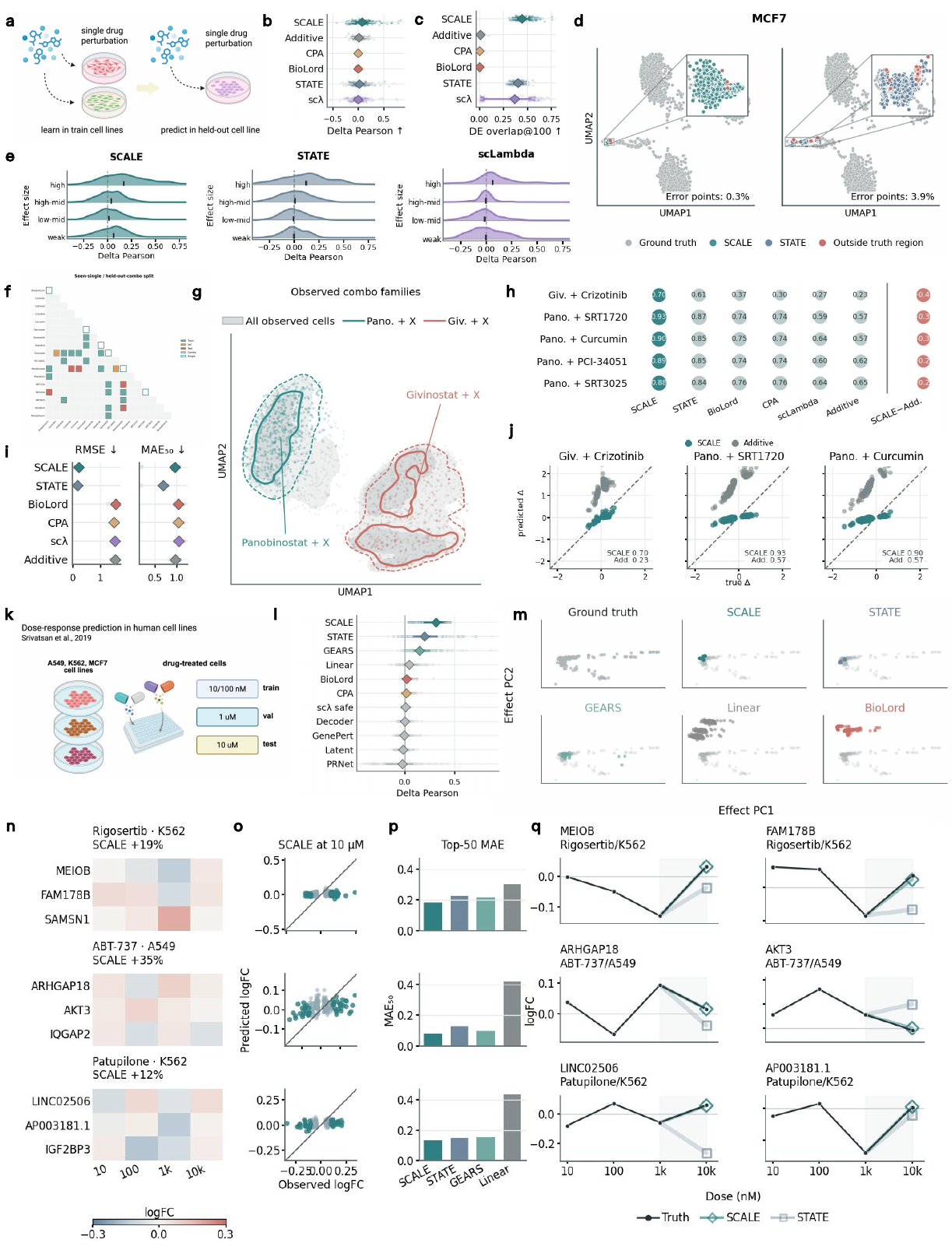}\par}
\par\vspace{1pt}
\scriptsize
\setstretch{0.96}
\justifying
\noindent\textbf{Drug response context transfer prediction (a--e).}
\textbf{(a)} Single-drug transfer from training cell lines to a held-out cell line.
\textbf{(b,c)} Condition-level $\Delta$ Pearson and top-100 differentially expressed gene overlap across models.
\textbf{(d)} Representative low-dimensional MCF7 response plot; red points fall outside the displayed observed-response region.
\textbf{(e)} $\Delta$ Pearson for SCALE, STATE and scLAMBDA across groups with different measured effect sizes.
\textbf{Combinations excluded from training (f--j).}
\textbf{(f)} Design that trains on single drugs and tests unseen combinations; this matrix does not define the separate five-pair analysis.
\textbf{(g)} UMAP of measured cells from the five-pair data, showing training, validation and test combinations with selected molecular structures; no predictions are shown.
\textbf{(h)} Pair-level $\Delta$ Pearson across six models and SCALE gain over direct addition.
\textbf{(i)} Overall RMSE and mean absolute error for the top 50 genes.
\textbf{(j)} Observed gene-level deltas compared with SCALE and direct-additive predictions for three representative pairs.
\textbf{High-dose prediction (k--q).}
\textbf{(k)} Split with 10 and 100~nM training doses, 1~$\mu$M validation and held-out 10~$\mu$M testing.
\textbf{(l)} Condition-level 10~$\mu$M $\Delta$ Pearson across models.
\textbf{(m)} Example low-dimensional response plot with mean absolute error and Spearman correlation for the top 50 genes.
\textbf{(n--p)} Measured effects by dose, SCALE predictions at 10~$\mu$M and mean absolute error for the top 50 genes in three selected drug--cell-line cases.
\textbf{(q)} Representative gene--dose trajectories; the shaded region marks the held-out 10~$\mu$M endpoint.
Each matched perturbation condition, rather than each cell or gene, is one unit in the overall comparisons. Low-dimensional plots are descriptive and do not prove that the full treated-cell distribution was recovered.
\par
\endgroup
\end{figure}

Drug-response prediction becomes more demanding when the cell line, drug pair or dose changes. A new cell line changes the cellular background. A held-out drug pair changes the treatment itself. A higher dose can change both the size and direction of the response. We tested SCALE under these three changes in the sci-Plex atlas\cite{srivatsan2020sciplex}. In every test, the model predicted the gene-expression change from a matched control population (Figure~\ref{fig:sciplex}).

First, SCALE outperformed the other evaluated models when transferring single-drug responses to a cell line excluded from training. This test asked whether a drug response learned in one cellular background remained useful in another. Across 2,244 matched conditions, SCALE achieved the highest median $\Delta$ Pearson(Figure~\ref{fig:sciplex}a,b). Its advantage was largest when the treatments produced stronger measured expression changes (Figure~\ref{fig:sciplex}e).

Correlation alone does not show whether the model identifies the same response genes. SCALE achieved the highest overlap between the predicted and measured top 100 differentially expressed genes (Figure~\ref{fig:sciplex}c). The MCF7 plot provided a separate visual check. Only 0.3\% of SCALE predictions fell outside the measured-response region, compared with 3.9\% for STATE (Figure~\ref{fig:sciplex}d). Together, the correlation and gene-overlap results show that SCALE carried drug-specific information into a new cellular context.

Second, SCALE recovered responses to all five drug pairs excluded from training and went beyond direct addition. A direct-addition baseline simply adds the responses to the two individual drugs. It therefore tests whether a drug pair contains information that is not captured by this simple sum. SCALE achieved the highest $\Delta$ Pearson for every held-out pair, with values from 0.70 to 0.93 (Figure~\ref{fig:sciplex}h). Its gain over direct addition ranged from 0.23 to 0.46. These results show that SCALE recovered combination-dependent response patterns that direct addition missed.

The measured-cell map explains the structure of this test. Combinations that shared panobinostat occupied one broad region, whereas givinostat plus crizotinib occupied another (Figure~\ref{fig:sciplex}g). Thus, the shared drug shaped the broad response, while the genes that distinguished each pair still had to be predicted. Because this panel contains measured cells rather than model predictions, model performance was assessed with the pair-level scores and gene-level comparisons in Figure~\ref{fig:sciplex}h,j.

The comparison with STATE also shows why the choice of metric matters. STATE had lower RMSE for four of the five pairs, whereas SCALE had higher $\Delta$ Pearson for all five (Figure~\ref{fig:sciplex}h,i). RMSE measures the numerical distance between prediction and measurement. By contrast, $\Delta$ Pearson measures whether genes follow the same response pattern. STATE was therefore numerically closer for several pairs, while SCALE better preserved the direction and relative ranking of gene changes within each pair. In three representative pairs, SCALE also followed the measured gene changes more closely than direct addition (Figure~\ref{fig:sciplex}j). These results support combination-dependent prediction, but they do not by themselves establish drug synergy.

Third, SCALE showed its clearest advantage at a 10~$\mu$M dose that was 100 times higher than the largest training dose. The model was trained on 10 and 100~nM, validated at 1~$\mu$M and tested only at 10~$\mu$M (Figure~\ref{fig:sciplex}k). The test dose was therefore 100-fold above the largest training dose and tenfold above the validation dose. At this unseen dose, SCALE achieved the highest mean $\Delta$ Pearson among the evaluated models.(Figure~\ref{fig:sciplex}l)

We next asked whether this advantage held across individual drug--cell-line conditions. Among 561 matched conditions, SCALE had lower top-50 mean absolute error than STATE in 94\% of comparisons. It also had higher top-50 Spearman correlation in 72\% and higher top-50 direction accuracy in 68\%. These metrics test three different properties: the size of the error, the ranking of the response genes and whether each gene increases or decreases. The gains across all three measures show that SCALE's advantage was not tied to one score.

Representative gene/dose trajectories explain what the condition-level gains mean. Rigosertib in K562, ABT-737 in A549 and patupilone in K562 included genes that changed direction or moved back towards baseline at 10~$\mu$M. In these examples, SCALE followed the measured high-dose endpoints more closely than STATE (Figure~\ref{fig:sciplex}n--q). It adjusted the prediction to each drug--cell-line setting instead of extending one common trend from the lower doses.

Together, the sci-Plex tests show that SCALE preserved drug-response specificity under three different changes. The cell-line test changed the cellular background, the pair test changed the treatment, and the dose test changed exposure. Across these settings, SCALE more consistently preserved which genes responded and how their changes were ordered. This extends the CRISPR analysis: Replogle showed that SCALE could separate target-specific responses from cell-line and average-treatment effects, while sci-Plex shows that distinct responses remain predictable when the cell context, drug partner or dose changes. Preserving these differences provides the information needed to compare candidate interventions before experimental testing.

\subsection{SCALE predicts unseen developmental perturbations and preserves stage-specific population structure}
\label{sec:developmental_generalization}

\begin{figure}[p]
\refstepcounter{figure}
\label{fig:developmental_generalization}
\addcontentsline{lof}{figure}{\protect\numberline{\thefigure}Developmental response recovery is constrained by experimental support and stage}
\begingroup
\footnotesize
\setstretch{1.0}
\noindent\textbf{Figure \thefigure\ | Developmental predictions depend on the amount of measured data and the developmental stage (a,b, task and split; c--g, accuracy and available data; h,i, genes excluded from training; j--n, cell types and time).}
\par\smallskip
{\centering
\includegraphics[width=0.871\textwidth]{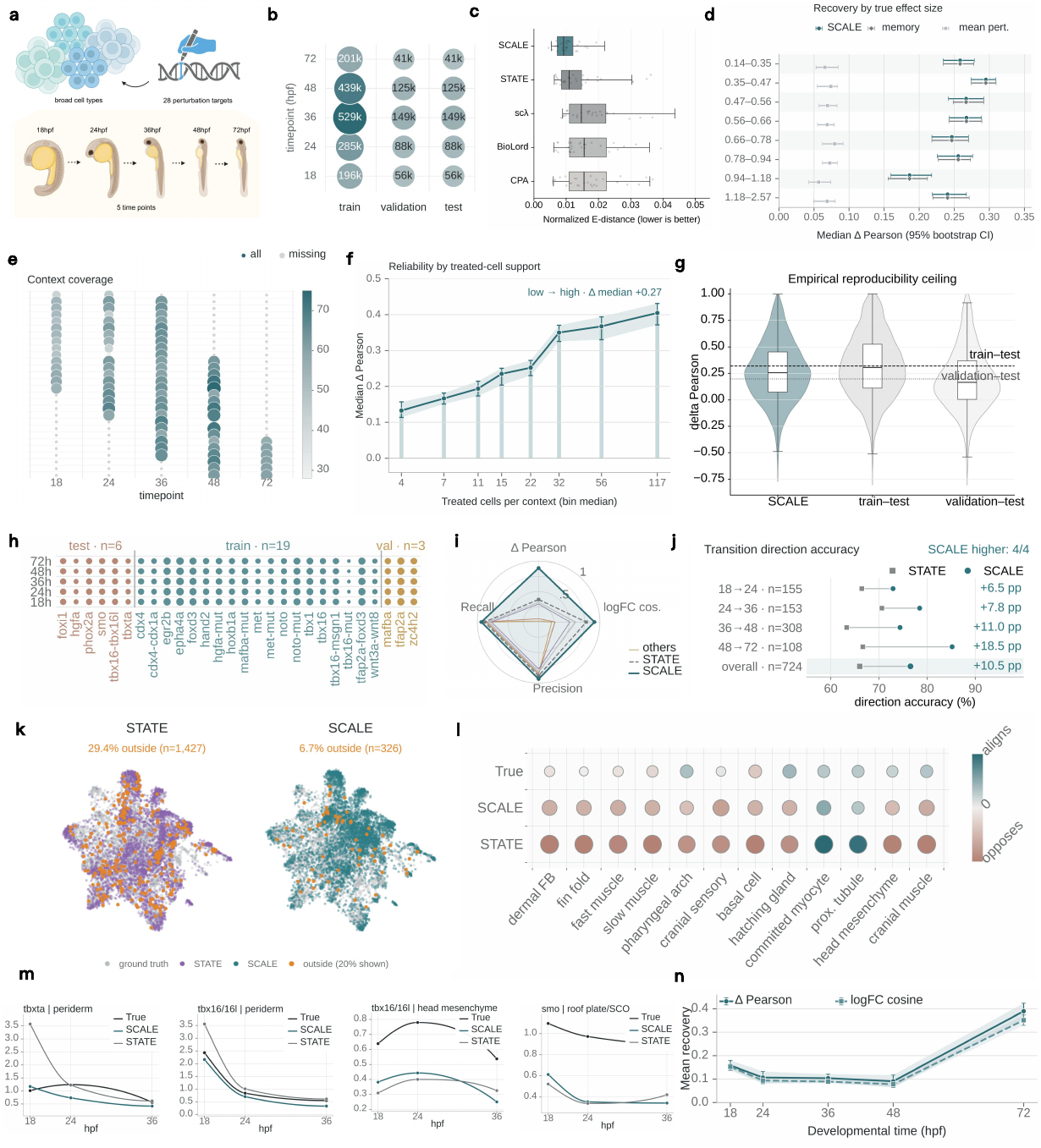}\par}
\par\vspace{1pt}
\scriptsize
\setstretch{0.98}
\justifying
\textbf{(a) Developmental perturbation atlas.} Twenty-eight genetic perturbations were measured across broad cell types at 18, 24, 36, 48 and 72 hpf.
\textbf{(b) Data split.} Cell numbers assigned to training, validation and test sets at each measured stage.
\textbf{(c) Overall effect accuracy.} Normalized E-distance across SCALE and external models; lower values indicate better predictions.
\textbf{(d) Accuracy by true effect size.} Median $\Delta$ Pearson and 95\% bootstrap confidence intervals across effect-size groups for SCALE and mean-response baselines.
\textbf{(e) Available cell contexts.} Measured and missing cell contexts at each stage; colour shows the number of measured contexts.
\textbf{(f) Accuracy by treated-cell count.} Median $\Delta$ Pearson across groups with different numbers of treated cells; error bars and shading show 95\% bootstrap confidence intervals, and stems mark group medians.
\textbf{(g) Measured reproducibility reference.} SCALE context-level $\Delta$ Pearson compared with train--test and validation--test agreement for repeated measurements of the same condition; dotted lines mark reference medians.
\textbf{(h) Genes excluded from training.} Six perturbations were used only for testing, with 19 genes used for training and three for validation; dots show the stages at which each target was measured.
\textbf{(i) Accuracy for genes excluded from training.} Normalized $\Delta$ Pearson, logFC cosine, DEG precision and recall for SCALE, STATE and external models.
\textbf{(j) Direction between stages.} Accuracy for four consecutive developmental changes and all changes combined; labels report the SCALE gain over STATE in percentage points.
\textbf{(k) Observed and predicted response locations.} Low-dimensional locations of observed and predicted responses; orange points mark predictions outside the region covered by observed responses, with 20\% shown for legibility.
\textbf{(l) Direction by cell type.} Cosine similarity between treatment effects and normal developmental change in matching controls, shown for measured responses, SCALE and STATE across well-measured broad cell types.
\textbf{(m) Selected changes over time.} Measured, SCALE-predicted and STATE-predicted effects for four test gene--cell-type settings at measured stages; connecting curves are visual guides only.
\textbf{(n) Accuracy by stage.} Mean SCALE $\Delta$ Pearson and logFC cosine across measured developmental stages, with 95\% hierarchical-bootstrap confidence intervals.

\noindent Teal, purple and grey denote SCALE, STATE and observed or reference values as specified in each panel. Contexts, rather than individual cells, are the units of evaluation.
\par
\endgroup
\end{figure}

The ZSCAPE benchmark contained 28 genetic perturbations measured across broad cell types at five developmental stages\cite{saunders2023embryo} (Figure~\ref{fig:developmental_generalization}a,b). For example, \textit{foxi1} was perturbed in embryos, and cells were collected at 18, 24, 36, 48 and 72 hpf. Here, \textit{foxi1} defines what was perturbed, and hpf defines when the response was measured. Beyond standard context-transfer evaluation, we tested gene-level out-of-distribution (OOD) generalization in a developmental setting. For a held-out gene such as \textit{foxi1}, all perturbed samples were placed in the test set, regardless of sampling time. The model had seen every developmental stage during training, but it had never seen the held-out gene perturbation. The test therefore asked whether SCALE could predict the effect of a new gene perturbation under a full developmental stages.

Under this unseen-gene split, SCALE achieved the lowest overall response error among the evaluated models against established models. SCALE had the lowest normalized E-distance, demonstrating the strongest overall agreement with measured responses on this metric (Figure~\ref{fig:developmental_generalization}c).

SCALE also generalized to six genetic perturbations excluded from both training and validation. We used 19 genes for training, three for validation and six only for testing (Figure~\ref{fig:developmental_generalization}h). Across the six test genes, SCALE achieved the strongest combined performance on $\Delta$ Pearson, logFC cosine, and DEG precision and recall (Figure~\ref{fig:developmental_generalization}i). Within the measured stages and observed cell types, SCALE therefore recovered both the direction of unseen responses and the genes involved.

SCALE preserved the direction of developmental responses across consecutive stages. SCALE consistently outperformed STATE across all developmental stage transitions, with an overall gain of 10.5 percentage points across all 724 transitions (Figure~\ref{fig:developmental_generalization}j). A complementary low-dimensional analysis showed that only 6.7\% of SCALE predictions fell outside the observed-response region, compared with 29.4\% for STATE (Figure~\ref{fig:developmental_generalization}k). Thus, SCALE preserved temporal response direction while keeping predictions within the observed-response region more consistently than STATE.

The cell-type map showed that developmental response directions remained cell-type specific: some perturbation responses aligned with normal developmental change, whereas others moved in the opposite direction (Figure~\ref{fig:developmental_generalization}l). SCALE captured these distinct patterns across four representative test gene--cell-type settings involving \textit{tbxta}, \textit{tbx16--tbx16l} and \textit{smo}, reproducing response shape or magnitude over time (Figure~\ref{fig:developmental_generalization}m). These examples connect the aggregate advantage to specific developmental trajectories rather than a single response shared across cell types.

Prediction accuracy varied sharply across developmental stages, showing a non-monotonic pattern where performance remained relatively modest through early stages before rising substantially at the latest stage (Figure~\ref{fig:developmental_generalization}n). This non-monotonic pattern shows that performance did not simply deteriorate as developmental time increased. Because all comparisons were made within measured stages and observed cell types, they do not establish generalization to absent stages or unmeasured cell types.

Finally, prediction accuracy was most strongly related to the number of treated cells within each gene/cell type/stage context. Accuracy increased substantially with greater numbers of treated cells, and this factor was a stronger predictor than developmental stage or effect size alone (Figure~\ref{fig:developmental_generalization}e,f).

Together, these results show that SCALE's population-level predictions retained gene-, stage- and cell-type-specific effects rather than relying on developmental stage alone. This extends the evidence from CRISPR and drug screens: an unseen perturbation can be distinguished by its predicted biological response across familiar developmental contexts.

\FloatBarrier

\subsection{SCALE generalizes cytokine responses across donors, cell types and patients}
\label{sec:pbmc_kang}

\begin{figure}[p]
\refstepcounter{figure}
\label{fig:pbmc_kang}
\addcontentsline{lof}{figure}{\protect\numberline{\thefigure}Immune-response prediction across donors, cell types and patients}
\begingroup
\footnotesize
\setstretch{1.0}
\noindent\textbf{Figure \thefigure\ | Immune-response prediction across donors, cell types and patients.}
\par\vspace{2pt}
\noindent\textit{Cytokine-atlas tests (a--f, h and i); patients excluded from training (j--l, n and o).}
\par\vspace{4pt}
{\centering
\makebox[\textwidth][c]{\includegraphics[width=0.99\textwidth]{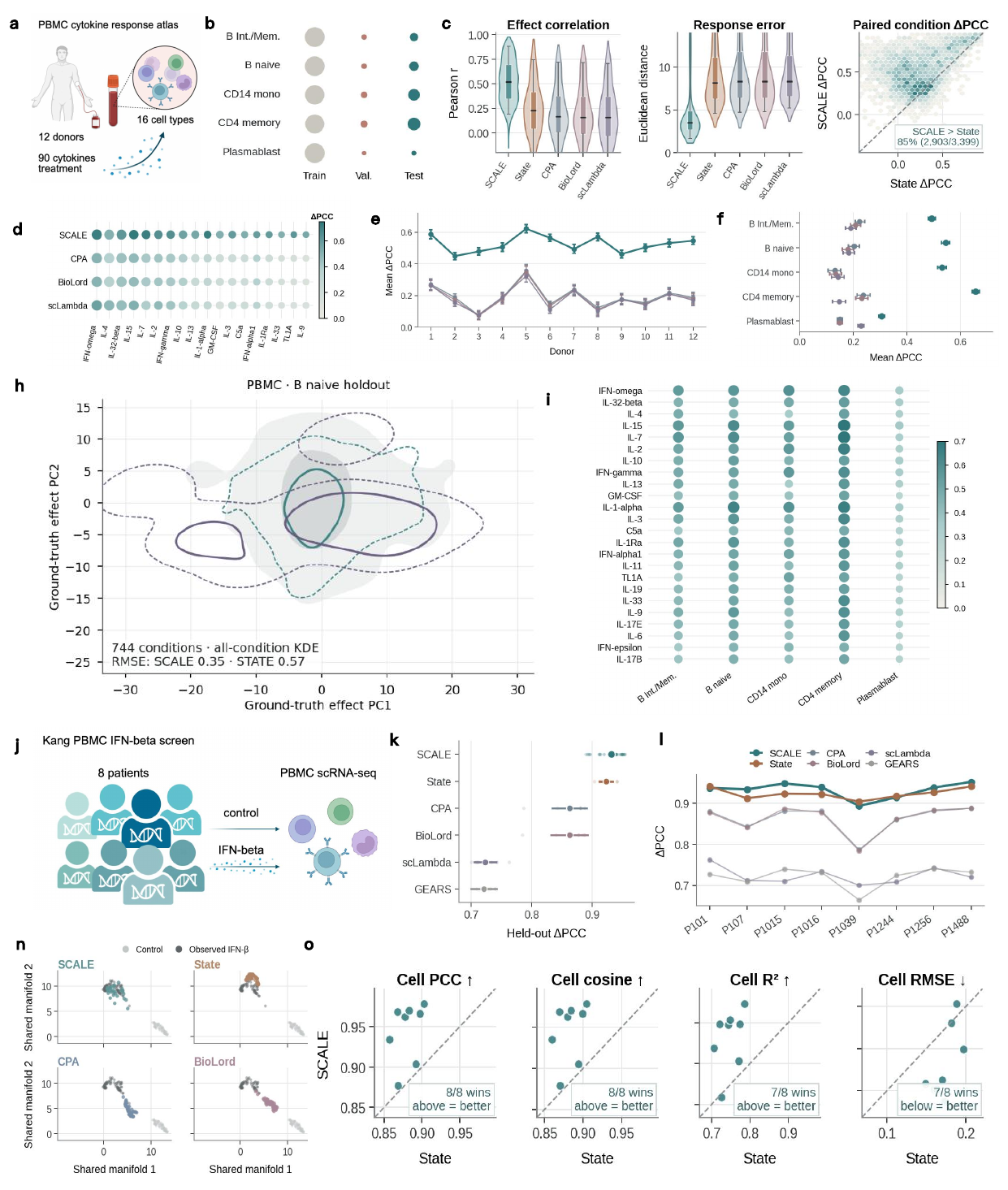}}\par}
\par\vspace{3pt}
\scriptsize
\setstretch{0.94}
\justifying
\noindent\textbf{(a) PBMC cytokine atlas.} Twelve donors, 90 cytokines and 16 cell types.
\textbf{(b) Cell types excluded from training.} Training, validation and test data for five excluded cell types.
\textbf{(c) Accuracy by condition.} Effect correlation, response error and paired SCALE--STATE $\Delta$PCC across 3,399 conditions.
\textbf{(d) Accuracy by cytokine.} Mean $\Delta$PCC for selected cytokines with strong effects.
\textbf{(e) Accuracy by donor.} Mean $\Delta$PCC across 12 donors.
\textbf{(f) Accuracy by cell type.} Mean $\Delta$PCC for each excluded cell type.
\textbf{(h) Response-space agreement in the B-naive holdout.} Kernel-density agreement with measured effects across 744 conditions; RMSE was 0.35 for SCALE and 0.57 for STATE.
\textbf{(i) DEG overlap.} Mean overlap of differentially expressed genes for each cytokine and cell type.
\textbf{(j) Kang patients excluded from training.} IFN-$\beta$ evaluation across eight patients.
\textbf{(k) Overall patient accuracy.} Test-patient $\Delta$PCC across six models.
\textbf{(l) Accuracy by patient.} Mean $\Delta$PCC for each test patient.
\textbf{(n) Low-dimensional response plot.} Predictions overlaid with control and observed IFN-$\beta$ cells.
\textbf{(o) Cell-level agreement.} Paired SCALE--STATE PCC, cosine similarity, $R^2$ and RMSE; one point per patient.
Each condition or test patient is one unit in the overall comparisons. Response-space projections in panels h and n are descriptive and do not prove that the full treated-cell distribution was recovered.
\par
\endgroup
\end{figure}

Cytokine-induced transcriptional responses vary across cell types and donors\cite{oesinghaus2025cytokine}. This context dependence raises a practical question: can a response be predicted when training excludes a donor-specific cytokine condition, an entire cell type or all cells from a patient? We tested the first two settings with the Human Cytokine Dictionary PBMC atlas and the patient-level setting with the independent Kang IFN-$\beta$ cohort\cite{kang2018multiplexed} (Figure~\ref{fig:pbmc_kang}). For the patient-level test, SCALE was trained and tested within the Kang cohort rather than transferred directly from the PBMC atlas.

With only part of each donor's cytokine panel included in training, SCALE predicted the held-out conditions more accurately than the compared models\cite{lotfollahi2023cpa,piran2024biolord,wang2024sclambda}. Across the PBMC atlas, it achieved the highest response correlation and the lowest response error (Figure~\ref{fig:pbmc_kang}c). Mean $\Delta$ Pearson was 0.525, and SCALE outperformed STATE in 85\% of paired conditions, showing that the advantage was not driven by a few cytokines (Figure~\ref{fig:pbmc_kang}c). The advantage remained across all 12 evaluation donors (Figure~\ref{fig:pbmc_kang}e). SCALE also recovered responses across interferons, interleukins and myeloid cytokines, showing that its performance was not limited to one cytokine family (Figure~\ref{fig:pbmc_kang}d). This setting therefore tested completion within partly measured donors rather than prediction for a new donor.

SCALE transferred cytokine responses to cell types left out of training. SCALE had the highest mean $\Delta$ Pearson in all five cell-type holdouts. Values ranged from 0.657 in CD4 memory T cells to 0.306 in plasmablasts (Figure~\ref{fig:pbmc_kang}b,f). In the B-naive holdout, SCALE more closely matched the measured response density, with a lower KDE RMSE than STATE (0.35 versus 0.57; Figure~\ref{fig:pbmc_kang}h). DEG overlap was highest in B-naive, CD14-monocyte and CD4-memory cells and lowest in plasmablasts (Figure~\ref{fig:pbmc_kang}i). Part of the cytokine response is therefore shared across cell types, but the exact genes and effect sizes still depend on cell type. 

SCALE also predicted IFN-$\beta$ responses for patients whose cells were completely absent from training. For each Kang split, one patient's control and stimulated cells were used only for testing. One patient was used for validation and the remaining six for training (Figure~\ref{fig:pbmc_kang}j). Across 56 test-patient and validation-patient combinations, mean $\Delta$PCC was 0.932 for SCALE and 0.924 for STATE. CPA and BioLORD each reached about 0.863 (Figure~\ref{fig:pbmc_kang}k). SCALE beat STATE for six of eight patient means, although both models were less accurate for patient P1039 (Figure~\ref{fig:pbmc_kang}l). In the displayed low-dimensional plot, SCALE predictions were closer to the measured IFN-$\beta$ region than predictions from the external models. This plot is a visual check rather than a test of full population agreement (Figure~\ref{fig:pbmc_kang}n). SCALE also beat STATE for cell-wise PCC and cosine similarity in all eight patients, and for $R^2$ and RMSE in seven of eight (Figure~\ref{fig:pbmc_kang}o). The small mean $\Delta$PCC gain shows more consistent prediction across patients, not superiority for every patient and metric.

Two complementary properties help explain these gains. First, SCALE's latent vectors retain response semantics that agree more closely with the measured endpoint. Here, ``semantic agreement'' does not mean visual separation alone: it means that nearby representations decode to similar cytokine-specific response directions, response genes and population distributions. The higher effect correlations, lower response errors, improved response-density agreement and DEG overlap provide gene- and population-level evidence for this interpretation (Figure~\ref{fig:pbmc_kang}c,h,i). The low-dimensional patient plot is consistent with it but, by itself, cannot establish semantic fidelity (Figure~\ref{fig:pbmc_kang}n).

Second, the set-aware encoder removes cell order as a nuisance variable while retaining population context. Its invariant summary satisfies $s(PX)=s(X)$ for any permutation $P$, and the shared cell update therefore satisfies $F(PX)=PF(X)$. Thus, shuffling the same cells cannot change the represented population or any order-aligned cell embedding. Mean aggregation also makes the pooled input to the population summary less sensitive to any one sampled cell; under independent sampling, its sampling error decreases at the usual $N^{-1/2}$ rate when the transformed cell features have finite variance, although robustness to rare-cell loss or distribution shift is not guaranteed by equivariance. This distinction is central: semantic fidelity concerns \emph{what biological response the embedding preserves}, whereas set equivariance concerns \emph{whether the representation changes under an irrelevant reordering}. The former is supported by agreement with measured responses; the latter follows exactly from the architecture. A stronger empirical robustness claim requires a separate stress test that repeatedly permutes, subsamples and perturbs the same cell population and compares embedding drift and prediction error with an otherwise matched cell-wise encoder.

The main cytokine response transferred more reliably than the exact response of each donor, cell type or patient. Across the three tests, SCALE recovered response programs that transferred across immune contexts without reducing all conditions to the same average response. Performance was weaker in plasmablasts and varied for some patients. The most consistent signals were cytokine-specific response patterns and response genes. These population-level predictions allowed us to compare cytokines by expected immune function rather than response magnitude alone. We used the predicted functional differences to select cytokines for donor-matched experiments.

\subsection{SCALE predictions of held-out cytokine responses prospectively guide PBMC experiments}
\label{sec:cytoti_welsh}
\label{sec:wetlab_validation}

\begin{figure}[p]
\refstepcounter{figure}
\label{fig:welsh_wetlab}
\addcontentsline{lof}{figure}{\protect\numberline{\thefigure}Prospective PBMC tests of four SCALE-predicted cytokine profiles}
\begingroup
\footnotesize
\setstretch{1.0}
\noindent\textbf{Figure \thefigure\ | Prospective donor-matched PBMC experiments recover four cytokine profiles selected from SCALE predictions.}
\par\vspace{2pt}
\noindent\textit{Prediction and study design (a); wet-lab readouts (b,c,e); cross-source agreement (d,f,g).}
\par\vspace{4pt}
{\centering
\makebox[\textwidth][c]{\includegraphics[width=0.90\textwidth]{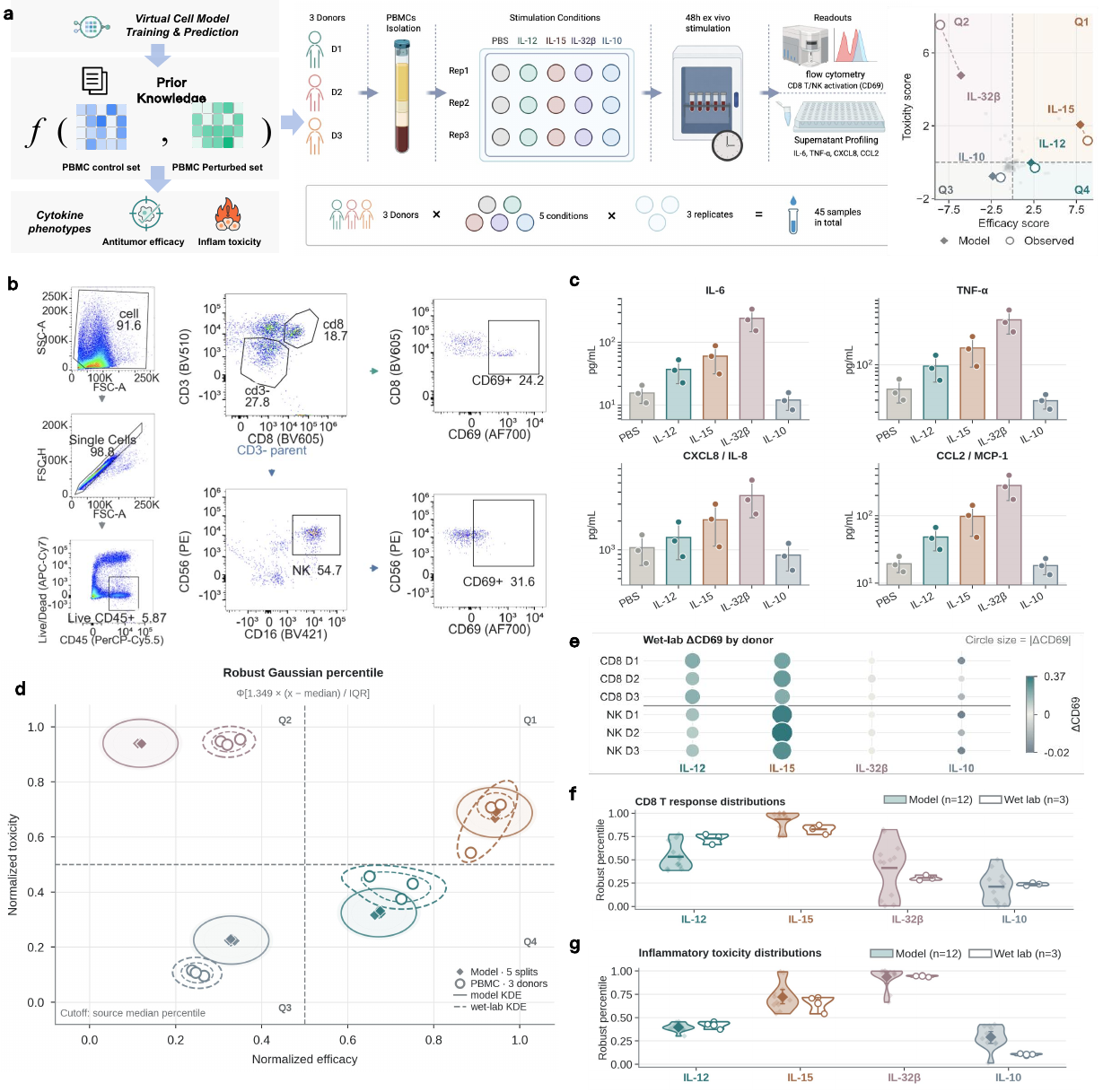}}\par}
\par\vspace{3pt}
\scriptsize
\setstretch{0.94}
\justifying
\noindent\textbf{(a) Prediction-to-experiment workflow.} SCALE combines the observed control population, perturbation and prior biological knowledge to predict cytokine phenotypes. PBMCs from three donors were treated with PBS, IL-12, IL-15, IL-32$\beta$ or IL-10 for 48~h with three technical replicates per donor--condition combination (45 culture wells), followed by CD69 flow cytometry and culture-liquid profiling.
\textbf{(b) Flow-cytometry cell selection.} The sequence identifies cells, single cells, live CD45-positive white blood cells, CD3-positive CD8-positive T cells and CD3-negative CD56/CD16-defined total NK cells before CD69 is measured.
\textbf{(c) Inflammatory secretion.} Donor-level concentrations are shown for IL-6, TNF-$\alpha$, CXCL8/IL-8 and CCL2/MCP-1 after each treatment. Each point is one donor after the three technical replicates are combined; bars summarize the three donors.
\textbf{(d) Joint efficacy--inflammation map.} Model results from five splits and PBMC measurements from three donors are placed on a common robust percentile scale, $\Phi[1.349(x-\mathrm{median})/\mathrm{IQR}]$, calculated separately within each source. Diamonds and solid contours show model results; open circles and dashed contours show wet-lab results. The median percentile defines the quadrant cutoffs. ``Toxicity'' on the plotted axis denotes the ex vivo inflammatory-secretion score, not in vivo toxicity.
\textbf{(e) Wet-lab CD69 by donor.} Matched-PBS changes in median CD69 are shown for CD8 T cells and total NK cells. Rows identify donor and cell type; circle colour shows the signed change and circle size its magnitude.
\textbf{(f) CD8 T-cell response distributions.} Robust-percentile distributions compare model predictions ($n=12$) with wet-lab donor measurements ($n=3$) for the four cytokines.
\textbf{(g) Inflammatory-secretion distributions.} The corresponding robust-percentile distributions compare predicted inflammation with the measured four-molecule culture-liquid score. Donor is the biological unit ($n=3$); technical replicates are combined before analysis.
\par
\endgroup
\end{figure}

To move from prediction to experiment, we asked whether SCALE could use distribution-level predictions for held-out cytokine responses to design an informative wet-lab test. SCALE was fitted without access to the target responses and then used to predict those cytokine conditions. No measured response from these test conditions, including the subsequent activation and supernatant outcomes, was available to the model or used for candidate selection. A single response-size score does not separate useful immune activation from inflammatory risk, so we applied the prespecified SCALE-CytoTI definitions directly to the predicted responses and selected cytokines by their predicted immune effects rather than by the largest expression changes. SCALE-CytoTI gives each predicted cytokine response two scores (Supplementary Fig.~\ref{fig:welsh_wetlab}). The predicted antitumour score adds effector-cell activation and antigen-presenting-cell (APC) priming, then subtracts suppressive responses. The predicted inflammation score uses gene-expression patterns in monocytes and dendritic cells that are linked to cytokine-release syndrome. Together, the scores form four groups with high or low predicted antitumour activity and inflammation. The line $\mathrm{CytoTI}=x-1.25y$ provides a secondary summary but does not replace the two scores.\par

The prediction-only map placed common cytokines into four groups with distinct profiles. IL-12 and type-I and type-II interferons had high predicted antitumour activity and low predicted inflammation. IL-15, IL-2 and IL-7 retained positive antitumour scores but had higher inflammation scores. IL-32$\beta$ and C5a had low antitumour scores and high inflammation scores, while IL-10 was low on both scores. These positions and group assignments were calculated from SCALE predictions before the wet-lab outcomes were available.\par

To test whether the predicted differences appeared in donor cells, we prospectively selected one cytokine from each group (Figure~\ref{fig:welsh_wetlab}a,d). The four cytokines were IL-12 (high antitumour score, low inflammation), IL-15 (high, high), IL-32$\beta$ (low, high) and IL-10 (low, low). The candidates and their group assignments were fixed from the SCALE predictions before any activation or supernatant measurements were inspected. This design compared IL-12 with IL-15 among cytokines with positive antitumour scores and IL-10 with IL-32$\beta$ among those with low scores. It also tested whether IL-12 and IL-15 activated CD8 T cells and NK cells more strongly than IL-32$\beta$ and IL-10.\par

We then tested the prespecified activation and inflammation predictions using new measurements that were unavailable at selection time. We predicted that IL-12 and IL-15 would increase CD69 in CD8 T cells and NK cells relative to matched PBS, whereas IL-32$\beta$ and IL-10 would cause weaker activation. We also predicted that IL-15 and IL-32$\beta$ would cause more secretion of inflammatory molecules than IL-12 and IL-10. These directional criteria were fixed before the experiment was analysed (Figure~\ref{fig:welsh_wetlab}a). Flow cytometry tested only the effector-cell component of the antitumour score because it measured CD8 T-cell and NK-cell activation, not APC priming or suppressive-cell responses. The figure uses ``inflammatory toxicity'' as a short label for the combined culture-liquid readout. This is an ex vivo measure of inflammatory secretion, not toxicity in an animal or patient.\par

The experiment compared five conditions in PBMCs from three independent donors. Cells were exposed for 48~h to PBS, IL-12, IL-15, IL-32$\beta$ or IL-10. Each donor--condition pair had three technical repeats, giving 45 culture wells in total (Figure~\ref{fig:welsh_wetlab}a). We analysed cells and matched culture liquid separately. Technical repeats were combined before comparison, and the donor was the biological unit ($n=3$). We therefore report each donor's response and whether donors agreed in direction. We do not treat individual wells as independent biological samples.

For the activation assay, flow cytometry measured CD69 separately in CD8 T cells and NK cells. The gating sequence selected cells, single cells and live CD45-positive white blood cells. CD8 T cells were CD3-positive and CD8-positive. Total NK cells were CD3-negative cells in the CD56/CD16-defined group (Figure~\ref{fig:welsh_wetlab}b). Donor-level changes in median CD69 relative to matched PBS are shown for both cell types in Figure~\ref{fig:welsh_wetlab}e. Figure~\ref{fig:welsh_wetlab}f separately compares the model-predicted and experimentally observed CD8 T-cell response distributions after each source is converted to a robust percentile scale.

Consistent with the predictions, IL-12 and IL-15 activated CD8 T cells and NK cells more strongly than IL-32$\beta$ and IL-10. After technical repeats were combined, all three donors showed increased CD69 after IL-12 and IL-15 in CD8 T cells, and IL-15 produced the largest NK-cell response across donors (Figure~\ref{fig:welsh_wetlab}e). IL-32$\beta$ remained near the matched PBS level in both cell types, while IL-10 was at or slightly below that level. The CD8 T-cell distributions retained the same broad ordering in the model and wet-lab data, with IL-15 highest, IL-12 next and IL-32$\beta$ and IL-10 lower (Figure~\ref{fig:welsh_wetlab}f). One CD8 T-cell measurement from donor D2 after IL-32$\beta$ had few recovered events and was interpreted cautiously. Together, these measurements supported the predicted separation between IL-12 and IL-15 and the two weakly activating cytokines.

The inflammatory-secretion assay produced a different ordering, with IL-32$\beta$ strongest, followed by IL-15, IL-12 and IL-10. We measured four inflammatory molecules in the matched culture liquid (Figure~\ref{fig:welsh_wetlab}c). IL-32$\beta$ caused the largest increase in all four across donors. Mean changes from matched PBS in $\log_{10}$ concentration were $+1.14$ for IL-6, $+1.01$ for TNF-$\alpha$, $+0.53$ for CXCL8 and $+1.12$ for CCL2. IL-15 caused intermediate increases ($+0.54$, $+0.59$, $+0.26$ and $+0.64$), while IL-12 caused smaller increases ($+0.32$, $+0.33$, $+0.10$ and $+0.37$). IL-10 remained near or below the matched baseline ($-0.12$, $-0.16$, $-0.08$ and $-0.05$). Absolute concentrations differed among donors, but the order remained the same. The model and wet-lab inflammatory distributions preserved this ordering after robust percentile normalization (Figure~\ref{fig:welsh_wetlab}g).

To compare model predictions and donor measurements without claiming equality of their raw scales, we normalized each source with its own median and interquartile range and converted the result to a robust Gaussian percentile. In the joint map, all four cytokines retained the same efficacy--inflammation quadrant across five model splits and three donors (Figure~\ref{fig:welsh_wetlab}d). The overlapping source-specific contours and the separate activation and inflammatory distributions (Figure~\ref{fig:welsh_wetlab}f,g) show agreement in relative profile and quadrant assignment, not numerical calibration of the two assays.

\FloatBarrier

Together, the two assays recovered the four cytokine profiles prospectively predicted by SCALE-CytoTI before the experimental outcomes were available. IL-12 activated CD8 T cells and NK cells with relatively little secretion of the measured inflammatory molecules. IL-15 combined strong activation, especially in NK cells, with more inflammatory secretion. IL-32$\beta$ caused the strongest inflammatory secretion without matching CD8 T-cell or NK-cell activation. IL-10 remained low in both measurements. All four cytokines retained their predicted efficacy--inflammation quadrant after source-specific robust normalization, and the separate activation and secretion distributions preserved the expected ordering. Thus, SCALE predictions for held-out target responses guided the selection of perturbations with distinct experimental profiles rather than merely ranking total response size.\par

The measured endpoints were CD69 activation and secretion of four inflammatory molecules in ex vivo PBMCs. APC priming, suppressive-cell responses, cellular sources of the secreted molecules and in vivo toxicity were outside this experiment. With three donors and four cytokines, the conclusion is limited to the response directions measured here.

\FloatBarrier
\section{Discussion}
\label{sec:discussion}

Single-cell perturbation experiments usually measure control and treated populations in different cells. This fact should define the prediction task. When true before--after cell pairs do not exist, forcing a model to learn such pairs makes the target depend on an artificial matching rule. The resulting predictions may look accurate only after averaging. A more direct task is to predict the measured treated population from the measured control population and perturbation.

The literature supports a narrower concern than the claim that distribution matching is itself flawed. Recent benchmarks show that average treatment effects and systematic shifts can dominate standard scores\cite{vinas2026systema,ahlmann2025linear}. Mode-collapse analyses attribute this behaviour to sparse perturbation signals, global reconstruction losses and permissive metrics\cite{mejia2025diversity,chi2026departures}. The risk therefore lies in weak condition-specific supervision and evaluation, not in the use of unpaired population endpoints itself.

Our results support this population-level view. They support two distinct claims. First, exact cell correspondence was not required to learn useful endpoint predictions. Second, SCALE's response representation did not collapse to one common effect in the tested settings. SCALE trains directly on measured control and treated populations. Its path through the learned representation is a calculation, not a measured biological trajectory. The main evidence is not only lower prediction error. SCALE also identifies perturbation-specific response direction, response genes and part of the treated population's range. In Replogle, median residual PCC was 0.33 for SCALE and 0.05 for STATE after average effects were removed. Direction@100 was 0.88 and 0.57, respectively, and only SCALE exceeded both mean baselines. Its latent responses retained target-specific direction and spread, whereas STATE occupied narrow regions. Drug, developmental and immune tests show that some of this information remains predictable for new cell lines, drug pairs, doses, genes, cell types and patients.

Average expression alone is an incomplete target. A prediction can match the group mean while losing the range of treated states and differences among cells. SCALE does not fully solve this problem. However, Replogle predictions preserve more of the relationships among gene perturbations than STATE. In sci-Plex, SCALE predictions more often move towards the measured treated region without the strongest compression or overshoot seen in some baselines. Developmental predictions more often remain within the measured response region. Immune predictions retain cytokine-response direction when the donor, cell type or patient changes. The low-dimensional plots provide descriptive evidence only. Together with the residual, directional and mean-baseline tests, these plots rule out a purely average-response explanation for SCALE on Replogle. They show that the learned endpoint map retained perturbation identity instead of merely fitting one shared distribution shift.

\noindent\textbf{\textit{Why population-delta learning matches unpaired endpoints.}}

The response variable learned by SCALE clarifies why this formulation is well matched to unpaired endpoint data. Let $S$ denote the set encoder, let $P_0$ and $P_1$ denote the observed control and perturbed populations, and define
\[
Z_0=S(P_0),\qquad Z_1=S(P_1),\qquad \Delta Z=Z_1-Z_0.
\]
SCALE models the baseline-centred conditional response distribution $p_\theta(\Delta Z\mid Z_0,c)$. For a fixed $Z_0$, the transformation $(Z_0,Z_1)\leftrightarrow(Z_0,\Delta Z)$ is one-to-one and has unit Jacobian because $Z_1=Z_0+\Delta Z$. Learning the delta is therefore information-equivalent to learning the endpoint; it does not create additional information, but places the prediction target in coordinates centred on change from the observed control population.

This change of coordinates also identifies a response that the experiment can determine without inventing cell pairs. A pointwise displacement $x_1-x_0$ depends on an unobserved coupling $\pi\in\Pi(P_0,P_1)$ between control and perturbed cells. By contrast, the set-level difference $S(P_1)-S(P_0)$ is fixed by the two observed populations and is unchanged by the choice of point-level coupling. Moreover, if $S(P)=\mathbb{E}_{X\sim P}[\phi(X)]$ uses a sufficiently informative feature map, then $S$ can distinguish distributions rather than only their raw means. Features that encode several orders and directions of variation allow $\Delta Z$ to represent coordinated changes in population structure. SCALE approximates this distribution-aware construction with a within-cell gene encoder, a permutation-invariant population summary and distribution matching in reconstruction\cite{zaheer2017deepsets,gretton2012kernel}. A finite learned embedding is not guaranteed to identify every possible distribution, so this argument specifies the intended statistical object rather than claiming perfect distribution recovery.

The delta factorization further removes components already present in the control population. If $Z_0=B+U$ and $Z_1=B+U+R(Z_0,c)+\varepsilon$, where $B+U$ is shared baseline structure and $R$ is the condition-dependent response, then
\[
\Delta Z=R(Z_0,c)+\varepsilon.
\]
The condition therefore needs to explain only the change added to the control representation. This is a recentering argument, not a claim that the delta has lower entropy than the endpoint. It also remains valid for the deployed endpoint objective. Defining $\widehat{\Delta Z}=\widehat Z_1-Z_0$ gives
\[
\lVert\widehat Z_1-Z_1\rVert_2^2
=\lVert\widehat{\Delta Z}-\Delta Z\rVert_2^2,
\]
so endpoint loss directly measures population-delta error. More formally, let $m(Z_0,c)=\mathbb{E}[\Delta Z\mid Z_0,c]$. Under squared loss, the condition-aware optimum is $f^*(Z_0,c)=m(Z_0,c)$, whereas a condition-blind model is restricted to $g^*(Z_0)=\mathbb{E}[\Delta Z\mid Z_0]$. Their minimum-risk difference is
\[
\mathcal{R}(g^*)-\mathcal{R}(f^*)
=\mathbb{E}\!\left[\left\lVert
m(Z_0,c)-\mathbb{E}[m(Z_0,c)\mid Z_0]
\right\rVert_2^2\right].
\]
This term is positive whenever the condition explains response variation that remains after the control population is fixed, thereby quantifying the cost of a condition-blind average response.

These properties define an inductive advantage rather than a strictly larger function class. Relative to CellFlow, which uses optimal transport to select a point-level coupling before learning cell-level velocities\cite{klein2025cellflow}, SCALE learns a coupling-independent response conditioned on a population representation. Relative to STATE\cite{adduri2025state}, the distinction is not stronger set expressivity: any endpoint map $F(Z_0,c)$ can be written as $Z_0+[F(Z_0,c)-Z_0]$. SCALE instead makes this population-delta factorization explicit, separating the input population, perturbation condition and predicted response. The theoretical advantage is therefore alignment between the learned response variable and the structure identifiable from destructive, unpaired experiments, rather than guaranteed dominance over every endpoint model.

The donor-matched PBMC experiment provides the strongest practical test. SCALE-CytoTI selected cytokines with different predicted immune effects instead of simply selecting the largest responses. Across three donors, the four cytokines produced different CD8 T-cell and NK-cell CD69 activation and different secretion of IL-6, TNF-$\alpha$, CXCL8 and CCL2. After source-specific robust normalization, all four cytokines retained the same efficacy--inflammation quadrant across five model splits and three donors. This experiment does not prove the numerical accuracy of CytoTI. It also does not test APC priming, suppressive cells or toxicity in animals or patients. It shows that the predicted differences were useful for choosing perturbations with distinct experimental profiles.

\noindent\textbf{\textit{Limitations and outlook.}}

The evidence has clear limits. Performance varies by dataset and metric, and no model is always best. STATE remains stronger on some Tahoe differential-expression metrics. Developmental accuracy depends strongly on the number of treated cells and the measured stage. SCALE should therefore not be viewed as a simulator of unmeasured development. The PBMC atlas includes some training cytokines from every test donor, so it predicts missing conditions for partly measured donors. It does not predict completely new donors. The Kang analysis tests new patients in a separate IFN-$\beta$ task and does not establish personalized clinical prediction.

These limits point to three next steps. First, endpoint-only training cannot reveal the biological events between control and treatment. Time-course or lineage-tracing experiments would be needed. Second, when control and treated populations are very similar, average-response corrections may still help. Future evaluations must continue to separate perturbation learning from donor, cell-type or cell-line averages. Third, several analyses predict response direction and gene ranking better than exact effect size. Matched experiments should test new doses, times, combinations and biological settings. More laboratory measurements are also needed to connect gene-expression predictions with cell function.

These observations do not establish that SCALE is immune to collapse. They show that collapse did not explain its predictions in the tested, experimentally supported settings. Collapse may still occur when condition labels are weak, response genes are sparse or a new context lies outside the training support.

In summary, SCALE separates two questions that are often conflated. Unpaired single-cell endpoints supported useful prediction without exact cell correspondence. Distribution-level transport also preserved target-specific directions, gene programmes and response geometry instead of collapsing to an average effect. These are empirical conclusions within the tested support, not a guarantee for arbitrary unseen biology. The cytokine experiment shows the practical value: this information can guide perturbations that produce different immune responses.

\section{Methods}
\subsection{Problem Formulation}
\label{sec:problem}

\noindent\textbf{SCALE predicts a treated cell population from a separately measured control population.}
Each training sample contains a control cell set $X_0$, a perturbed cell set $X_1$, and observed conditions including cell type $c$, perturbation identity $p$, and experimental batch $b$:
\begin{equation}
\mathcal{D}=\{(X_0, X_1, c, p, b)\}.
\end{equation}
Here $X_0, X_1 \in \mathbb{R}^{B \times N \times G}$ denote batches of cell sets, where $B$ is the batch size, $N$ is the number of cells sampled per example, and $G$ is the number of genes. Within each example, cells are treated as an unordered set rather than a sequence.

The same cell cannot be measured both before and after treatment. SCALE therefore does not seek a one-to-one cell match. Instead, it learns how treatment changes the distribution of a control population. Formally, we seek to learn the conditional law
\begin{equation}
\hat{X}_1 \sim p_\theta(X_1 \mid X_0, c, p, b),
\end{equation}
where the target is the treated population given the control population and the observed conditions.

This target makes the task a population transport problem rather than cell-wise regression. The model must change the population according to the observed conditions while retaining meaningful differences among cells. SCALE performs this task in a learned, lower-dimensional latent space, which reduces the difficulty of working directly with sparse, high-dimensional expression data (Supplementary Fig.~\ref{fig:model}).

\subsection{A Hierarchical Set-Aware Latent Cell-State Encoder}
\label{sec:latent_encoder}

\noindent\textbf{The encoder first represents each cell and then adds information from the whole population.}
Each example is an unordered set of cells, and each cell contains a high-dimensional gene-expression profile. We therefore first model gene relationships within each cell and then summarize information across cells.

\noindent\textbf{The same gene encoder represents every cell independently.}
Let $x_i \in \mathbb{R}^{G}$ denote the expression vector of the $i$-th cell in a population. We first apply a shared gene-wise encoder
\begin{equation}
f_{\mathrm{gene}} : \mathbb{R}^{G} \rightarrow \mathbb{R}^{d},
\end{equation}
to each cell independently, yielding
\begin{equation}
h_i = f_{\mathrm{gene}}(x_i), \qquad i = 1,\dots,N.
\end{equation}
In our implementation, $f_{\mathrm{gene}}$ is a LLaMA-style attention module\cite{touvron2023llama} or a basic attention module that operates on gene features \emph{within} one cell. Cells do not interact at this stage: the same encoder is applied independently to every cell. The output is therefore permutation equivariant, meaning that reordering the input cells produces the same reordering of the output embeddings:
\begin{equation}
F_{\mathrm{gene}}(PX) = P F_{\mathrm{gene}}(X),
\end{equation}
where \(X \in \mathbb{R}^{B \times N \times G}\) is a batch of cell sets, \(P\) reorders the cell dimension, and \(F_{\mathrm{gene}}(X)\) returns cell-wise embeddings. For the \(b\)-th cell set, let
\begin{equation}
H^{(b)} = \left[h^{(b)}_1,\ldots,h^{(b)}_N\right] \in \mathbb{R}^{N \times d_h},
\end{equation}
where \(h^{(b)}_i \in \mathbb{R}^{d_h}\) denotes the gene-level embedding of the \(i\)-th cell. For notational
simplicity, we omit the batch index \(b\) below and write \(h_i \in \mathbb{R}^{d_h}\).

\noindent\textbf{DeepSets adds population-wide information without depending on cell order.}
The cell embeddings are next processed by a DeepSets layer\cite{zaheer2017deepsets} across the cell dimension. We compute a permutation-invariant summary, meaning that its value does not change when cells are reordered:
\begin{equation}
s(X) = \rho\!\left(\frac{1}{N}\sum_{i=1}^{N}\phi(h_i)\right) \in \mathbb{R}^{d_s},
\end{equation}
where \(\phi : \mathbb{R}^{d_h} \rightarrow \mathbb{R}^{d_\phi}\) and
\(\rho : \mathbb{R}^{d_\phi} \rightarrow \mathbb{R}^{d_s}\) are learnable mappings. In batched form,
stacking the set-level summaries across the \(B\) samples yields
\(S(X) \in \mathbb{R}^{B \times d_s}\).

This summary is then fed back to refine each cell representation through a learnable fusion map
\begin{equation}
z_i = \psi\!\left(h_i, s(X)\right), \qquad
\psi : \mathbb{R}^{d_h} \times \mathbb{R}^{d_s} \rightarrow \mathbb{R}^{d}, \qquad
i = 1,\ldots,N,
\end{equation}
so that \(z_i \in \mathbb{R}^{d}\). Stacking the refined cell embeddings gives the final latent population
\begin{equation}
Z = \left[z_1,\ldots,z_N\right] \in \mathbb{R}^{N \times d},
\end{equation}
and, in batched form, \(Z \in \mathbb{R}^{B \times N \times d}\).

By construction, \(s(X)\) does not change when cells are reordered, while the refined cell representations follow the new order. The encoder can therefore add population-wide information to each cell without treating cell order as meaningful.

\noindent\textbf{The final representation contains both within-cell gene information and population context.}
The LLaMA-style gene encoder models gene relationships within each cell, whereas DeepSets summarizes the population across cells. The resulting \textbf{\emph{set-aware latent population representation}} therefore describes both individual cells and the unordered population to which they belong.

\noindent\textbf{The autoencoder is trained to reconstruct both individual cells and the overall population.}
Let the full encoder and decoder be denoted by
\begin{equation}
Enc_\phi : \mathbb{R}^{B \times N \times G} \rightarrow \mathbb{R}^{B \times N \times d},
\qquad
Dec_\psi : \mathbb{R}^{B \times N \times d} \rightarrow \mathbb{R}^{B \times N \times G},
\end{equation}
where $Enc_\phi$ is the hierarchical encoder described above. It processes a batch of $B$ cell sets, each containing $N$ cells and $G$ genes per cell.

Applied to the control and perturbed populations, we obtain
\begin{equation}
Z_0 = Enc_\phi(X_0), \qquad Z_1 = Enc_\phi(X_1), \qquad
Z_0, Z_1 \in \mathbb{R}^{B\times N\times d},
\end{equation}
where $X_0, X_1 \in \mathbb{R}^{B\times N\times G}$.

The decoder reconstructs expression cell-wise,
\begin{equation}
\hat{X}_0 = Dec_\psi(Z_0),\qquad
\hat{X}_1 = Dec_\psi(Z_1), \qquad
\hat{X}_0, \hat{X}_1 \in \mathbb{R}^{B\times N\times G}.
\end{equation}

We train the latent space with a joint MSE-MMD objective\cite{gretton2012kernel}. MSE measures reconstruction error for individual expression values, whereas MMD measures the difference between the reconstructed and observed population distributions:
\begin{equation}
\mathcal{L}_{\mathrm{AE}}
=
\mathcal{L}_{\mathrm{MSE}}
+
\lambda_{\mathrm{MMD}} \mathcal{L}_{\mathrm{MMD}},
\end{equation}
where
\begin{align}
\mathcal{L}_{\mathrm{MSE}}
&=
\mathbb{E}_{X_0}
\left[
\left\|
\hat{X}_0 - X_0
\right\|_2^2
\right]
+
\mathbb{E}_{X_1}
\left[
\left\|
\hat{X}_1 - X_1
\right\|_2^2
\right], \\
\mathcal{L}_{\mathrm{MMD}}
&=
\mathbb{E}\!\left[
\mathrm{MMD}^2(\hat{X}_0, X_0)
\right]
+
\mathbb{E}\!\left[
\mathrm{MMD}^2(\hat{X}_1, X_1)
\right].
\end{align}
Here, $\lambda_{\mathrm{MMD}}$ balances cell-level reconstruction and population-level distribution matching. The MMD term is
\begin{equation}
\mathrm{MMD}^2(A,B)
=
\mathbb{E}_{a,a' \sim A}[k(a,a')]
+
\mathbb{E}_{b,b' \sim B}[k(b,b')]
-
2\mathbb{E}_{a \sim A,\, b \sim B}[k(a,b)],
\end{equation}
where $k(\cdot,\cdot)$ is a positive-definite kernel, such as a Gaussian kernel.

\subsection{Conditional Latent Flow Transport under Population Endpoint Supervision}

\noindent\textbf{Training predicts the measured treated endpoint and does not claim to recover an unobserved biological trajectory.}
The training data contain corresponding control and treated populations, but no before--after cell pairs or intermediate measurements. We therefore formulate prediction as conditional transport between two observed endpoints in latent space. We use a construction related to flow matching\cite{lipman2023flowmatching}, but do not interpret its interpolation path as biological time. Let
\begin{equation}
Z_0 = Enc_{\phi}(X_0), \qquad Z_1 = Enc_{\phi}(X_1), \qquad Z_0, Z_1 \in \mathbb{R}^{B \times N \times d}
\end{equation}
denote the latent populations encoded from the control and treated cell sets. We connect them with a linear training path
\begin{equation}
Z_t = (1 - t) Z_0 + t Z_1, \qquad t \sim \mathrm{Uniform}(0, 1),
\end{equation}
This interpolation is only a training device. It is not a model of biological dynamics, and matching row positions in the two sampled sets are not observed before--after cell pairs. Here, \(t\) is a scalar sampled uniformly from the unit interval.

Given the interpolated latent distribution \(Z_t\), time \(t\), and observed conditions \(C\), the conditional transport backbone produces hidden latent states
\begin{equation}
H_t = g_{\theta}(Z_t, t, C), \qquad H_t \in \mathbb{R}^{B \times N \times d}.
\end{equation}
Here, \(g_{\theta}(\cdot)\) is the conditional transport model with parameters \(\theta\). The final {\ProjectName} model directly predicts the treated endpoint:
\begin{equation}
\hat{Z}_1 = h_x(H_t),
\end{equation}
where \(h_x(\cdot)\) is the endpoint prediction head and \(\hat{Z}_1\) is the predicted treated population in latent space. The model is trained with
\begin{equation}
\mathcal{L}_{\mathrm{flow}} = \mathcal{L}_x = \mathbb{E}_t \left[ \left\| \hat{Z}_1 - Z_1 \right\|_2^2 \right].
\end{equation}
Here, \(\mathbb{E}_t[\cdot]\) averages over sampled values of \(t\). This is the JiT-inspired x-pred/x-loss strategy\cite{li2025back}. It trains the model against the measured treated endpoint, which is the information available in the data.

\noindent\textbf{Endpoint prediction with endpoint loss performed best among the tested JiT variants.}
For the ablation study, we tested a broader family of JiT variants\cite{li2025back}. They predict either the treated endpoint or the change from control to treatment, and they can be trained against either target. This comparison tests how the prediction target and loss affect learning from matched control and treated populations.

Let the endpoint and displacement heads be denoted by
\begin{equation}
\hat{Z}_1 = h_x(H_t),
\end{equation}
\begin{equation}
\hat{U} = h_v(H_t),
\end{equation}
where \(h_v(\cdot)\) predicts the change in latent space and \(\hat{U}\) is the predicted change. The two outputs are related by
\begin{equation}
\hat{Z}_1 = Z_0 + \hat{U}, \qquad \hat{U} = \hat{Z}_1 - Z_0.
\end{equation}

We further define the endpoint and displacement supervision spaces abstractly as
\begin{equation}
\mathcal{L}_x = \mathbb{E}_t \left[ \left\| \hat{Z}_1 - Z_1 \right\|_2^2 \right],
\end{equation}
\begin{equation}
\mathcal{L}_v = \mathbb{E}_t \left[ \left\| \hat{U} - U^{\star} \right\|_2^2 \right],
\end{equation}
where \(U^{\star} = Z_1 - Z_0\) is the difference between the sampled latent endpoint arrays. This tensor difference is a training target, not an observed change in an individual cell. These definitions yield four JiT variants:
\begin{align}
\textit{x-pred / x-loss:} \qquad & \hat{Z}_1 = h_x(H_t), && \mathcal{L} = \mathcal{L}_x,\\
\textit{x-pred / v-loss:} \qquad & \hat{Z}_1 = h_x(H_t), \ \hat{U} = \hat{Z}_1 - Z_0, && \mathcal{L} = \mathcal{L}_v,\\
\textit{v-pred / x-loss:} \qquad & \hat{U} = h_v(H_t), \ \hat{Z}_1 = Z_0 + \hat{U}, && \mathcal{L} = \mathcal{L}_x,\\
\textit{v-pred / v-loss:} \qquad & \hat{U} = h_v(H_t), && \mathcal{L} = \mathcal{L}_v.
\end{align}

The JiT variants substantially outperform the base flow formulation. Among the completed configurations, direct endpoint prediction trained with endpoint loss gives the strongest results. We therefore use x-pred/x-loss as the default {\ProjectName} configuration and report the full comparison in the ablation study.

\noindent\textbf{The final objective matches the measurements available during training.}
The model learns conditional transport between measured endpoints rather than reconstructing an unobserved random process. Its training target therefore matches the endpoint used for evaluation.

\subsection{Modality-aware condition representation and injection}

\noindent\textbf{Each type of experimental condition is encoded in a form suited to that data type.}
SCALE does not force category labels, molecular features and continuous measurements into the same raw format. Each observed variable $a_k$ is processed by an encoder $f_k$ suited to its type, and the resulting tokens are then mapped to a common width:
\begin{equation}
e_k=f_k(a_k)\in\mathbb{R}^{d_c},\qquad
C=[e_1;\ldots;e_K]\in\mathbb{R}^{K\times d_c}.
\end{equation}
This design lets each input keep its original meaning while giving the transport model a common input format.

\noindent\textbf{Category labels use learned embeddings.}
For PBMC and Replogle, perturbation identity, cell type, donor, batch and cell line are categorical variables. An index or equivalent one-hot vector selects an entry from a learned embedding table, followed by a multilayer perceptron (MLP):
\begin{equation}
e_k=\operatorname{MLP}_k(E_k[a_k]),
\end{equation}
where $E_k\in\mathbb{R}^{K_k\times d_k}$. The one-hot vector is therefore only a way to select the correct embedding; it does not imply that all conditions belong to one shared one-hot space.

\noindent\textbf{Drug structure and dose are encoded together.}
For Tahoe-100M, the perturbation token combines a 2,048-dimensional molecular fingerprint $r_p$ with the continuous concentration $s$. The fingerprint is projected and normalized. Concentration receives a sinusoidal encoding and then changes the molecular representation through feature-wise linear modulation:
\begin{align}
h_p &= \operatorname{LN}(W_r r_p),\\
[\gamma(s),\beta(s)] &= \operatorname{MLP}_s(\operatorname{PE}(s)),\\
e_p &= \operatorname{MLP}_p\!\left(\gamma(s)\odot h_p+\beta(s)\right).
\end{align}
Cell-line and plate variables keep their categorical encoders. This branch represents molecular similarity and dose dependence, which a drug label alone cannot capture.

\noindent\textbf{Gene features and developmental time allow the model to represent genes left out of training.}
For the developmental experiment, a continuous gene-target fingerprint $r_g$ represents the held-out target instead of a target-ID lookup. Developmental time is represented by both a learned stage embedding and a sinusoidal encoding of measured hours after fertilization. We combine the two time representations with the gene representation:
\begin{align}
e_g &= f_g(r_g),\\
e_\tau &= f_\tau\!\left([E_\tau[\tau];\operatorname{PE}(\tau)]\right),\\
e_{g\times\tau} &= f_{g\times\tau}\!\left([e_g;e_\tau;e_g\odot e_\tau]\right).
\end{align}
The gene, time, gene--time interaction and cell-type tokens are combined before transport. A held-out gene can therefore enter the condition encoder through its fingerprint without needing a learned embedding that was trained for that specific gene.

\noindent\textbf{Attention combines all condition tokens and supplies them to every transport block.}
The type-specific tokens are projected to the transport width and combined by learned attention. With a seed query $q_{\mathrm{seed}}$, we compute
\begin{align}
\tilde C &= \Pi(C),\\
\alpha &= \operatorname{Softmax}\!\left(\frac{(q_{\mathrm{seed}}W_Q)(\tilde C W_K)^\top}{\sqrt d}\right),\\
c_{\mathrm{cond}} &= \alpha(\tilde C W_V).
\end{align}
The condition summary is combined with the sinusoidal encoding of interpolation time and appended to the latent cell tokens in every transport block. If $H_\ell\in\mathbb{R}^{B\times N\times d}$ denotes the cell tokens, the joint update is applied to $[H_\ell;c_{\mathrm{cond}}]$, after which only the first $N$ outputs are retained. The cell dimension has no positional encoding, and the same token-wise parameters are used for all cells. Adding condition information therefore does not make the model depend on cell order.

\subsection{Generation in expression space}

\noindent\textbf{At prediction time, SCALE encodes control cells, predicts the treated representation and decodes gene expression.}
The input is a control set $X_{0}$ with conditions $(c,p,b)$. The encoder computes $Z_{0} = Enc_{\phi}(X_{0})$, and the trained transport model predicts $\hat{Z}_{1}$. The decoder maps this predicted endpoint to
\begin{equation}
\hat{X}_{1} = Dec_{\psi}(\hat{Z}_{1}).
\end{equation}
This produces a deterministic prediction of the treated population in gene-expression space for the observed biological context.

\subsection{Training-support prior-delta centering}

\noindent\textbf{Prior-delta centering adjusts only the predicted population mean and keeps cell-to-cell differences unchanged.}
The final prediction step includes prior-delta centering (PDC). PDC adjusts the decoded population using treatment effects estimated only from the training split. Let $u$ denote the available biological context, $p$ a non-control perturbation and $0$ the matched control. For the raw decoded cells $\{\hat{x}_{u,p,i}\}_{i=1}^{N}$, define the model-predicted group mean
\begin{equation}
\hat{\mu}_{u,p}=\frac{1}{N}\sum_{i=1}^{N}\hat{x}_{u,p,i}.
\end{equation}
Let $d^{\mathrm{train}}_{u,p}$ be the change from control to treatment estimated from training data, and let $\hat{\mu}_{u,0}$ be the model-predicted control mean. The training-supported target mean is
\begin{equation}
m^{\mathrm{prior}}_{u,p}=\hat{\mu}_{u,0}+d^{\mathrm{train}}_{u,p}.
\end{equation}
PDC shifts every predicted cell in the group by the same vector:
\begin{equation}
\tilde{x}_{u,p,i}=\hat{x}_{u,p,i}
+\lambda\left(m^{\mathrm{prior}}_{u,p}-\hat{\mu}_{u,p}\right),
\qquad 0\leq\lambda\leq1.
\end{equation}
The corrected mean is therefore $(1-\lambda)\hat{\mu}_{u,p}+\lambda m^{\mathrm{prior}}_{u,p}$, while each cell's difference from the population mean remains unchanged:
\begin{equation}
\tilde{x}_{u,p,i}-\tilde{\mu}_{u,p}
=\hat{x}_{u,p,i}-\hat{\mu}_{u,p}.
\end{equation}
PDC therefore adjusts the population-level change without replacing the cell-to-cell variation generated by the model. When the exact context--perturbation effect is available in training, PDC uses it. Otherwise, a fixed training-only lookup uses the same perturbation in other contexts, the overall training effect or zero, in that order. The adjustment weight and fallback rule are selected on validation data and then fixed for test evaluation. Test responses are never used to construct the target means.

\subsection{Benchmark preprocessing, splits and evaluation}

\noindent\textbf{All model comparisons use matched preprocessing, split definitions and evaluation rules.}
For PBMC and Tahoe-100M, raw counts are normalized by each cell's total counts, transformed with $\log(1+x)$, rescaled by a factor of 10 and restricted to the top 2,000 highly variable genes using Scanpy\cite{wolf2018scanpy}, following the benchmark protocol\cite{yuan2026perturbdiff}. Replogle-Nadig follows the processed STATE release, including filtering for effective target knockdown, normalization by total counts, log transformation, constant rescaling and restriction to the shared 2,000 genes\cite{adduri2025state,yuan2026perturbdiff}. The sci-Plex analyses use the processed Srivatsan profiles with matched vehicle controls. Treatment effects are calculated within each cell line and dose before comparison across contexts\cite{srivatsan2020sciplex}.

PBMC\cite{oesinghaus2025cytokine} reserves four of 12 donors for evaluation. For these donors, 30\% of perturbations are shown during training and 70\% are evaluated. This is therefore completion within partly observed donors, not strict transfer to entirely unseen donors. Tahoe-100M\cite{zhang2025tahoe100m} holds out five biologically different cell lines and does not use data from those test lines during model development. Replogle-Nadig\cite{replogle2022mapping,nadig2025trade} holds out one cell line but shows the model some perturbations from that line during training. The sci-Plex evaluation has three separate tests: transfer to a cell line excluded from its training fold; prediction of five drug pairs left out of training using the same observed and predicted files for every model; and dose extrapolation, with 10 and 100~nM used for training, 1~$\mu$M for validation and 10~$\mu$M for testing. A separately generated design that trains on single drugs and tests unseen combinations contains a partly different set of drug pairs, so we do not treat it as the same split. Kang\cite{kang2018multiplexed} is a separate, cohort-specific task. Each fold reserves one patient's control and IFN-$\beta$-treated cells for testing, one patient for validation and the other six for training. It does not test direct transfer from a model trained on the Human Cytokine Dictionary PBMC atlas. The developmental tests use ZSCAPE\cite{saunders2023embryo} and remove each test gene perturbation completely from training.

Cell-Eval metrics\cite{adduri2025state,yuan2026perturbdiff} are calculated from exported cell-level predictions. Predicted and observed responses use the same control populations. In Table~\ref{tab:table1_unified_benchmark}, we align published metric names as follows: PR-Recall with DE Precise, FC Spearman with DE Spearman fold change and DE Overlap Accuracy with DE Overlap. Rows marked as reproduced were generated in our implementation. All unmarked baseline values are reported at the precision given by PerturbDiff and retain their published source.

\subsection{SCALE-CytoTI scoring and Welsh experimental-validation design}

\noindent\textbf{SCALE-CytoTI separates predicted antitumour activity from predicted inflammation.}
For each cytokine $c$, SCALE-CytoTI summarizes the predicted response using standardized scores for activation of tumour-killing immune cells, priming of antigen-presenting cells (APCs), suppressive immune responses and monocyte/dendritic-cell inflammation linked to cytokine-release syndrome. The two plotted coordinates are
\begin{align}
E_c &= z_c(\mathrm{effector})+z_c(\mathrm{APC\ priming})-z_c(\mathrm{suppression}),\\
T_c &= z_c(\mathrm{monocyte/DC\ CRS\mbox{-}like\ inflammation}),
\end{align}
where $E_c$ is predicted antitumour efficacy and $T_c$ is predicted inflammatory toxicity. Zero on each standardized axis divides the map into four groups. The dashed decision boundary is
\begin{equation}
\mathrm{CytoTI}_c=E_c-1.25T_c=0.
\end{equation}
The Welsh map is used to select candidates. The PBMC map uses the same definitions but serves only as a reference. It is not a separate experimental replicate or validation cohort.

Cytokines were chosen to cover all four combinations of high or low efficacy and toxicity: IL-12 (high efficacy, low toxicity), IL-15 (high efficacy, high toxicity), IL-32$\beta$ (low efficacy, high toxicity) and IL-10 (low efficacy, low toxicity). The choices were fixed from the predicted map before the activation and supernatant measurements were inspected.

PBMCs from three independent donors were treated for 48~h with PBS, IL-12, IL-15, IL-32$\beta$ or IL-10. Each donor--condition combination had three technical replicates, giving 45 culture wells. Flow cytometry identified viable CD8 T cells and natural killer (NK) cells. CD69 measured activation, and CD56 and CD16 defined the NK-cell group. Matching culture supernatants were measured for IL-6, TNF-$\alpha$, CXCL8 and CCL2. Technical replicates were combined within each donor and condition, and changes were calculated relative to matching PBS controls. Donors, not individual wells or cell events, were treated as biological replicates ($n=3$).

Because the Welsh study contains only three independent donors, we focus on the size and direction of each change from PBS and on consistency across donors. We do not calculate statistical significance from technical replicates. The efficacy prediction is supported only if high-efficacy cytokines produce greater CD8 T-cell or NK-cell activation than low-efficacy cytokines. The inflammatory-risk prediction is supported only if high-risk cytokines cause greater secretion of IL-6, TNF-$\alpha$, CXCL8 and CCL2 than the lower-risk cytokines used for comparison.

\bibliography{main}

\appendix
\appendix
\clearpage

\section{Computational phenotypes underlying SCALE-CytoTI screening}

\noindent\textbf{SCALE-CytoTI uses matched endpoint responses to define computational phenotypes for cytokine screening.}
For each donor $d$, annotated cell type $t$, cytokine $c$ and gene $g$, the input to phenotype scoring is a prespecified standardized contrast between the predicted cytokine-treated endpoint and its matched phosphate-buffered saline (PBS) pseudobulk profile,
\begin{equation}
\Delta_{d,t,c,g}=\mathcal{S}\!\left(\widehat{x}_{d,t,c,g},x^{\mathrm{PBS}}_{d,t,g}\right),
\end{equation}
where $\mathcal{S}$ is fixed before evaluation and may be implemented as a log fold change or another standardized endpoint difference. Matching within donor and cell type removes baseline differences that would otherwise dominate absolute expression. The resulting quantities describe predicted cytokine-induced changes; they are computational phenotypes used to rank candidates, not observed clinical phenotypes.

\noindent\textbf{The effector phenotype represents cytotoxic lymphocyte activation.}
It is evaluated mainly in CD8 T, NK, NKT-like and other cytotoxic T-cell populations. NKG7, GNLY, GZMB and PRF1 form the core granule-mediated killing programme, with IFNG providing a complementary type-1 effector signal\cite{wen2022nkg7,pena1997granulysin,packard2007granzyme}. CCL5, XCL1 and XCL2 receive lower weights because they connect cytotoxic lymphocytes to recruitment of cross-presenting cDC1 cells\cite{bottcher2018cdc1}. KLRD1 and FCGR3A are treated as lineage- and receptor-context features rather than universal positive effectors; CX3CR1 primarily marks differentiated effector states, and CST7 is excluded from the positive core or assigned a negative weight because cystatin F can reduce granzyme maturation and cytotoxicity\cite{masilamani2006nkg2a,bottcher2015cx3cr1,prunk2020cystatinf}.

\noindent\textbf{The APC-priming phenotype separates antigen presentation from a generic interferon response.}
It is calculated only in annotated monocytes, dendritic cells, plasmacytoid dendritic cells and B cells. HLA-DRA, HLA-DRB1, HLA-DPA1 and CD74 represent MHC-II presentation, whereas B2M, TAP1 and TAPBP represent MHC-I peptide transport and loading\cite{koeffler1984hlad,cresswell1999mhc1}. CD80, CD86, CD40, CD83 and IL12B capture costimulation, maturation, CD40 licensing and IL-12-supported T-cell priming\cite{lanier1995cd80cd86,mackey1998cd40,snijders1998il12,prechtel2007cd83}. CXCL9 and CXCL10 contribute an interferon-responsive lymphocyte-recruitment component but are not interpreted as direct measures of antigen processing\cite{lim2024cxcl9cxcl10}.

\noindent\textbf{The suppression phenotype combines Treg-like and suppressive-myeloid programmes.}
Within CD4/Treg populations, FOXP3 and CTLA4 receive the greatest weights, supported by IL2RA, TIGIT, LAG3, IL10 and TGFB1\cite{hori2003foxp3,wing2008ctla4,chinen2016il2r,joller2014tigit}. IKZF2 and TNFRSF18 receive lower weights because Helios is more informative as a stability marker than as an essential suppressive driver, whereas GITR stimulation can weaken tolerance\cite{lam2022helios,shimizu2002gitr}. Within myeloid populations, S100A8, S100A9, ARG1, IL1RN, VEGFA, MMP9, CXCL8 and CD274 capture MDSC-like or suppressive states. ARG1 and S100A8/S100A9 have the strongest direct support; CD274 and CXCL8 remain context dependent because they can also reflect interferon feedback or acute inflammation\cite{cheng2008s100a9,steggerda2017arginase,lu2016pdl1,alfaro2016il8}.

\noindent\textbf{The inflammatory-toxicity phenotype captures a CRS-like myeloid response rather than clinical toxicity itself.}
It is evaluated mainly in CD14 and CD16 monocytes, dendritic cells and related myeloid populations. IL6, TNF, IL1B, CXCL8, CCL2, CCL3, CCL4, CSF2, NFKBIA, NLRP3, PTGS2, SOD2 and SERPINE1 represent cytokine production, NF-$\kappa$B activation, inflammasome activity, oxidative stress and endothelial or coagulation injury. IL6, IL1B, CCL2 and CSF2 receive the greatest weights because monocyte- and macrophage-derived IL-1 and IL-6 mediate cytokine-release syndrome, early CCL2 associates with severe responses and GM-CSF inhibition reduces monocyte-dependent inflammatory cytokine production\cite{giavridis2018crs,norelli2018crs,hay2017crs,sterner2019gmcsf,sachdeva2019gmcsf}. NLRP3 contributes to a broader inflammasome programme, whereas NFKBIA, PTGS2, SOD2 and SERPINE1 are auxiliary stress or feedback markers and are not sufficient alone to define inflammatory risk\cite{liu2020pyroptosis}.

\noindent\textbf{Module scores are aggregated into a safety-first cytokine ranking.}
For module $m$, cell-type-level responses are combined as a fixed weighted score,
\begin{equation}
S^{(m)}_{d,t,c}=\sum_{g\in G_m}w_{m,g}\Delta_{d,t,c,g},
\end{equation}
then aggregated across the prespecified cell populations and standardized across cytokines. This yields effector, APC-priming, suppression and inflammatory-toxicity scores. The general therapeutic-index form is
\begin{equation}
\mathrm{CytoTI}_c=z_c(\mathrm{effector})+z_c(\mathrm{APC\ priming})-z_c(\mathrm{suppression})-\lambda z_c(\mathrm{toxicity}).
\end{equation}
The reported maps retain the two coordinates separately as $E_c=z_c(\mathrm{effector})+z_c(\mathrm{APC\ priming})-z_c(\mathrm{suppression})$ and $T_c=z_c(\mathrm{toxicity})$, and use $\lambda=1.25$ only for the dashed secondary decision boundary. In screening, very high inflammatory toxicity is considered before efficacy, followed by high suppression; high effector or APC-priming scores are labelled favourable only when inflammatory toxicity is not high. Weak or donor-inconsistent responses are retained as context dependent rather than called universally active.

\noindent\textbf{Weights, thresholds and donor aggregation are fixed without access to held-out outcomes.}
The gene sets provide biological anchors, but their exact weights, $\lambda$ and decision thresholds are modeling choices rather than a standardized clinical index. They must be estimated with training data and fixed before evaluation on held-out donors or cytokines. Donor stability is assessed separately using the median response, the fraction of donors changing in the same direction and between-donor dispersion. Donor-disjoint and cytokine-disjoint evaluations are therefore needed to test whether the computational phenotype transfers beyond the samples used to define it. When SCALE outputs endpoint expression, the same matched-PBS rules can be applied directly; a downstream classifier operating on latent representations must instead be trained only to reproduce labels derived from measured endpoint data.

\FloatBarrier

\section{Supplementary model architecture}

\begin{figure}[H]
    \centering
    \includegraphics[width=0.9\textwidth]{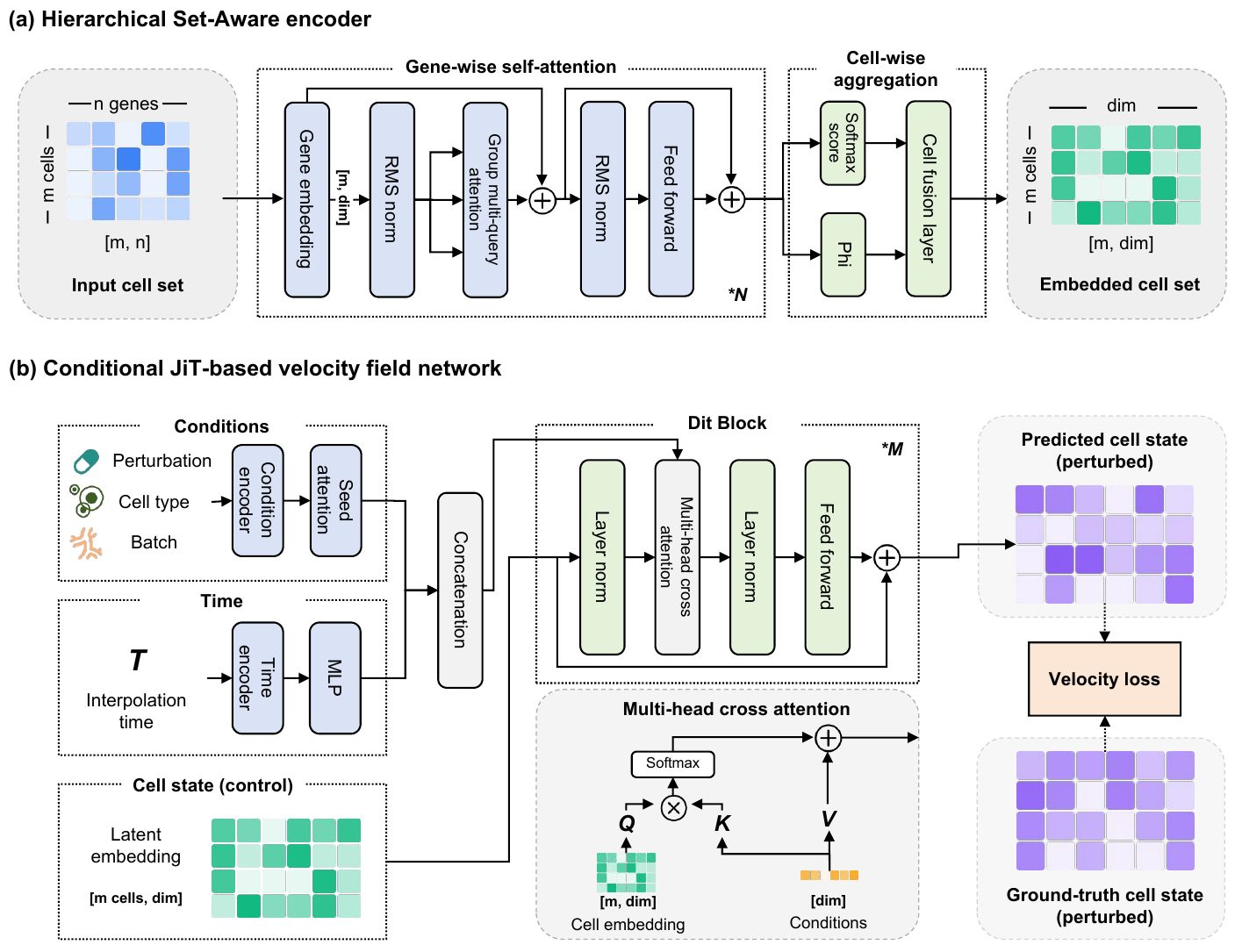}
    \caption{\textbf{Supplementary \ProjectName architecture (a, set-aware encoder; b, conditional JiT transport).} (a) \textbf{Set-aware encoder.} Gene-expression profiles are converted into cell embeddings and combined with a population summary that does not depend on cell order. (b) \textbf{Conditional JiT transport network.} Experimental conditions, interpolation time and cell embeddings are combined to predict the treated endpoint in latent space.}
    \label{fig:model}
\end{figure}

\noindent\textbf{SCALE encodes cells together with their population, predicts the treated endpoint and decodes it back to gene expression.}
Supplementary Fig.~\ref{fig:model} shows the set-aware encoder and conditional transport network in detail. The model combines experimental conditions, interpolation time and cell embeddings to predict the treated endpoint in latent space. The decoder then maps that prediction back to gene-expression space.

\section{Perturbation Metrics}
\label{app:metric_def}
\label{app:context_experiment_metrics}

\noindent\textbf{The evaluation tests average expression, treatment-induced changes and recovery of differentially expressed genes.}
We use the protocol from CellFlow~\cite{klein2025cellflow} and Cell-Eval version 0.6.6 from STATE. Because sequencing destroys the measured cells and cell states vary naturally, predicted cells cannot be matched one to one with measured cells. We therefore compare statistics calculated across populations.

For clarity, the evaluation metrics are organized along two axes.

\noindent\textbf{Population-expression metrics compare summaries of predicted and measured cell populations.}
These metrics test whether the predicted expression values match the corresponding measured population.

\noindent\textbf{Gene-level metrics test whether predicted responses recover the measured pattern of gene changes.}
These metrics compare differentially expressed genes, their direction of change and their ranking.

For a fair comparison, the same control cells are used to calculate predicted and measured responses. Differences in performance therefore come from the predicted treated cells.

\noindent\textbf{We compare each prediction with the measured treated population from the same condition.}
Let $M$ denote the total number of perturbation conditions.
For a perturbation $\lambda$, let
$\{\mathbf{x}_{\lambda,i}^{\text{pert}}\}_{i=1}^{N^{\lambda}_{\text{pert}}}$ and
$\{\mathbf{x}_{\lambda,i}^{\text{ctrl}}\}_{i=1}^{N^{\lambda}_{\text{ctrl}}}$
denote the ground-truth expression profiles of perturbed and control cells.
Let
$\{\hat{\mathbf{x}}_{\lambda,i}^{\text{pert}}\}_{i=1}^{N^{\lambda}_{\text{pert}}}$
denote the predicted perturbed cells.
In practice, Cell-Eval uses
$N^{\lambda}_{\text{ctrl}} = N^{\lambda}_{\text{pert}}$.

We define pseudobulk expression as the mean expression across cells:

\begin{equation}
\bar{\mathbf{x}}_{\lambda}^{\text{pert}}
=
\frac{1}{N^{\lambda}_{\text{pert}}}
\sum_{i=1}^{N^{\lambda}_{\text{pert}}}
\mathbf{x}_{\lambda,i}^{\text{pert}},
\qquad
\bar{\mathbf{x}}_{\lambda}^{\text{ctrl}}
=
\frac{1}{N^{\lambda}_{\text{ctrl}}}
\sum_{i=1}^{N^{\lambda}_{\text{ctrl}}}
\mathbf{x}_{\lambda,i}^{\text{ctrl}} .
\end{equation}

The pseudobulk profile for predicted perturbed cells is

\begin{equation}
\bar{\hat{\mathbf{x}}}_{\lambda}^{\text{pert}}
=
\frac{1}{N^{\lambda}_{\text{pert}}}
\sum_{i=1}^{N^{\lambda}_{\text{pert}}}
\hat{\mathbf{x}}_{\lambda,i}^{\text{pert}} .
\end{equation}

The \textbf{relative perturbation effect} is the difference between treated and control pseudobulk profiles:

\begin{equation}
\Delta \mathbf{x}_{\lambda}
=
\bar{\mathbf{x}}_{\lambda}^{\text{pert}}
-
\bar{\mathbf{x}}_{\lambda}^{\text{ctrl}},
\end{equation}

\begin{equation}
\Delta \hat{\mathbf{x}}_{\lambda}
=
\bar{\hat{\mathbf{x}}}_{\lambda}^{\text{pert}}
-
\bar{\mathbf{x}}_{\lambda}^{\text{ctrl}}.
\end{equation}

These quantities are used in the metrics below.

\noindent\textbf{These metrics test whether predicted average expression matches the measured average.}
\label{app:metric_accuracy}

\begin{itemize}

\item \textbf{Coefficient of Determination (R$^2$).}
The $R^2$ metric measures how much of the variation in the measured data is explained by the predictions, relative to a mean-value baseline:
\[
R^2 = 1 - \frac{\sum_i (y_i - \hat{y}_i)^2}{\sum_i (y_i - \bar{y})^2}.
\]
Higher values indicate better predictive accuracy.

\item \textbf{Perturbation Discrimination Score (PDS).}
PDS tests whether each predicted response is closer to its matching measured response than to responses from other perturbations:
\[
\text{PDS} = 1 - \frac{1}{M}\sum_{\lambda=1}^{M}\frac{r_{\lambda}}{M},
\quad
r_{\lambda} =
\sum_{p \neq \lambda}
\mathbbm{1}\!\left[
d(\Delta \hat{\mathbf{x}}_{\lambda}, \Delta \mathbf{x}_p)
<
d(\Delta \hat{\mathbf{x}}_{\lambda}, \Delta \mathbf{x}_{\lambda})
\right].
\]

A value of $1$ indicates perfect discrimination, while a value close to $0.5$ corresponds to random performance.
We report three variants using different distance functions:
L1 distance (PDS$_{\mathrm{L1}}$), L2 distance (PDS$_{\mathrm{L2}}$), and cosine distance (PDS$_{\mathrm{cos}}$).

\item \textbf{Pearson Delta Correlation (PDCorr).}
For each perturbation $\lambda$, we calculate the Pearson correlation between predicted and measured treatment effects:
\[
\text{PDCorr} =
\frac{1}{M}
\sum_{\lambda=1}^{M}
\mathrm{PearsonR}
(\Delta \hat{\mathbf{x}}_{\lambda}, \Delta \mathbf{x}_{\lambda}).
\]
This statistic is also called Pearson $\Delta$ and is commonly used in single-cell perturbation-prediction benchmarks~\cite{roohani2024gears,adduri2025state,vinas2026systema}.

\item \textbf{Mean Absolute Error (MAE).}
MAE measures the absolute difference between predicted and measured treatment effects:
\[
\mathrm{MAE} =
\frac{1}{M}
\sum_{\lambda=1}^{M}
\left\|
\Delta \hat{\mathbf{x}}_{\lambda} -
\Delta \mathbf{x}_{\lambda}
\right\|_1.
\]

\item \textbf{Mean Squared Error (MSE).}
MSE gives more weight to large errors by using squared Euclidean distance:
\[
\mathrm{MSE} =
\frac{1}{M}
\sum_{\lambda=1}^{M}
\left\|
\Delta \hat{\mathbf{x}}_{\lambda} -
\Delta \mathbf{x}_{\lambda}
\right\|_2^2.
\]

\end{itemize}

\subsection{Replogle cross-context response diagnostics}
\label{app:replogle_metrics}

These quantities are statistical prediction diagnostics rather than direct biological measurements. The condition-level response PCC uses the same basic statistic as Pearson $\Delta$ or PDCorr: a Pearson correlation between predicted and measured treatment-effect vectors~\cite{roohani2024gears,adduri2025state,vinas2026systema}. ``Response PCC'' is a descriptive label for the condition-level score used here; Pearson $\Delta$ and PDCorr are the names more commonly used in perturbation-prediction studies. In the Replogle analysis, the vectors contain signed log$_2$ fold changes rather than the pseudobulk differences defined above. The exact names and formulas for shared-response PCC, context-modulation PCC and response-component error are study-specific and are defined below. Earlier studies motivate separating shared or systematic effects from context-dependent perturbation responses, but do not define these three diagnostics~\cite{vinas2026systema,xu2025cellcap}.

Let $\mathbf{r}_{p,c}\in\mathbb{R}^{2000}$ and $\hat{\mathbf{r}}_{p,c}\in\mathbb{R}^{2000}$ denote the measured and predicted signed log$_2$ fold-change vectors for target $p$ in cell line $c$. Both vectors use the same 2,000 HVGs and the matched non-targeting control for that cell line. Fold changes are clipped below at $10^{-8}$ before applying $\log_2$, and the signed log$_2$ fold-change entry is set to zero for a gene absent from an input differential-expression table.

\paragraph{Condition-level response PCC.}
For each target--cell-line condition, response PCC is
\begin{equation}
R_{p,c}
=
\operatorname{corr}_{g}
\left(\hat{\mathbf{r}}_{p,c},\mathbf{r}_{p,c}\right),
\end{equation}
where the correlation is calculated across the 2,000 genes. Each target--cell-line condition is one evaluation unit. The paired SCALE--STATE comparison contained 4,058 conditions.

\paragraph{Shared response and context modulation.}
Let $\mathcal{C}_p$ be the set of cell lines available for target $p$. We included a target only when measured, SCALE and STATE responses were all available in at least three cell lines. This gave 738 targets: 544 measured in three cell lines and 194 measured in four. Cell lines receive equal weight. The response shared across cell lines is
\begin{equation}
\mathbf{s}_p
=
\frac{1}{|\mathcal{C}_p|}
\sum_{c\in\mathcal{C}_p}\mathbf{r}_{p,c},
\qquad
\hat{\mathbf{s}}_p
=
\frac{1}{|\mathcal{C}_p|}
\sum_{c\in\mathcal{C}_p}\hat{\mathbf{r}}_{p,c}.
\end{equation}
The cell-line-specific deviations from this shared response are
\begin{equation}
\mathbf{m}_{p,c}=\mathbf{r}_{p,c}-\mathbf{s}_p,
\qquad
\hat{\mathbf{m}}_{p,c}=\hat{\mathbf{r}}_{p,c}-\hat{\mathbf{s}}_p.
\end{equation}
We then define
\begin{align}
R^{\mathrm{shared}}_p
&=
\operatorname{corr}_{g}
\left(\hat{\mathbf{s}}_p,\mathbf{s}_p\right),\\
R^{\mathrm{mod}}_p
&=
\operatorname{corr}_{c,g}
\left(
\operatorname{vec}(\hat{\mathbf{m}}_{p,c}),
\operatorname{vec}(\mathbf{m}_{p,c})
\right).
\end{align}
These are the shared-response PCC and context-modulation PCC, respectively. Each target contributes one value to each metric, and the Results report the median across the 738 targets.

\paragraph{Response-component error.}
For each target, shared-response strength and the context-modulation index are
\begin{align}
A_p &= \|\mathbf{s}_p\|_2,\\
I_p
&=
\frac{
|\mathcal{C}_p|^{-1}
\sum_{c\in\mathcal{C}_p}\|\mathbf{m}_{p,c}\|_2
}{A_p+10^{-12}},
\end{align}
with $\hat A_p$ and $\hat I_p$ defined in the same way for the prediction. The response-component error is
\begin{equation}
E_p
=
|\hat A_p-A_p|
+
|\hat I_p-I_p|.
\end{equation}
Lower values indicate closer agreement. The Results report the raw $E_p$, whereas Figure~\ref{fig:replogle}h plots $\log(1+E_p)$ to compress the long upper tail. Because the first term is an absolute difference in a 2,000-gene log-fold-change norm and the second is a difference between ratios, this study-specific summary is used only to compare models under the same Replogle preprocessing and gene set; it should not be compared numerically across datasets.

For every PCC above, non-finite gene pairs are removed. A correlation is undefined when fewer than three finite pairs remain or when either vector has a standard deviation below $10^{-12}$; undefined values are excluded from summaries. These diagnostics describe prediction agreement, not statistical significance or biological mechanism. Pearson correlation also does not measure response magnitude and can reward systematic shifts shared across perturbations, so we interpret it together with the residual, mean-prior, direction and response-component analyses~\cite{vinas2026systema}.

\noindent\textbf{These metrics test whether the model identifies the genes, directions and rankings that change after treatment.}
\label{app:context_experiment_metrics_depattern}

Cell-Eval identifies differentially expressed (DE) genes with the Wilcoxon rank-sum test. Because thousands of genes are tested, $p$-values are adjusted with the Benjamini--Hochberg procedure to control the false discovery rate (FDR). DE analysis is performed separately on measured and predicted cells.

\noindent\textbf{We define differentially expressed genes separately for the measured and predicted responses.}

Let $\mathcal{G}$ be the top 2,000 highly variable genes (HVGs) used for prediction. For perturbation $\lambda$, a gene $g \in \mathcal{G}$ is called differentially expressed when its adjusted $p$-value satisfies $p_{\mathrm{adj}} < 0.05$.

Let $\mathcal{G}^{\mathrm{DE}}_{\lambda}$ be the set of significant DE genes in the measured response and $\hat{\mathcal{G}}^{\mathrm{DE}}_{\lambda}$ the corresponding set from predicted cells.

Genes are ranked by absolute log-fold change:
\[
|\log \mathrm{FC}_{\lambda,g}| =
\left|
\log_2
\frac{
\bar{\mathbf{x}}^{\text{pert}}_{\lambda,g} + \epsilon
}{
\bar{\mathbf{x}}^{\text{ctrl}}_{\lambda,g} + \epsilon
}
\right|.
\]

Predicted fold changes are calculated in the same way from predicted pseudobulk expression. Let $\mathcal{G}^{k}_{\lambda}$ and $\hat{\mathcal{G}}^{k}_{\lambda}$ be the top-$k$ ranked DE genes from measured and predicted cells.

\begin{itemize}

\item \textbf{DE Overlap (DEOver).}
\[
\mathrm{DEOver}_k =
\frac{1}{M}
\sum_{\lambda=1}^{M}
\frac{
|\mathcal{G}^{k}_{\lambda} \cap \hat{\mathcal{G}}^{k}_{\lambda}|
}{k}.
\]
Here $k = |\mathcal{G}^{\mathrm{DE}}_{\lambda}|$.

\item \textbf{DE Precision (DEPrec).}
\[
\mathrm{DEPrec}_k =
\frac{1}{M}
\sum_{\lambda=1}^{M}
\frac{
|\mathcal{G}^{k}_{\lambda} \cap \hat{\mathcal{G}}^{k}_{\lambda}|
}{
|\hat{\mathcal{G}}^{k}_{\lambda}|
}.
\]
Here $k = |\hat{\mathcal{G}}^{\mathrm{DE}}_{\lambda}|$.

\item \textbf{Direction Agreement (DirAgr).}
Let
\[
\mathcal{G}^{\cap}_{\lambda}
=
\hat{\mathcal{G}}^{\mathrm{DE}}_{\lambda}
\cap
\mathcal{G}^{\mathrm{DE}}_{\lambda}.
\]

\[
\mathrm{DirAgr} =
\frac{1}{M}
\sum_{\lambda=1}^{M}
\frac{
|
\{g \in \mathcal{G}^{\cap}_{\lambda} :
\mathrm{sgn}(\widehat{\log \mathrm{FC}}_{\lambda,g}) =
\mathrm{sgn}(\log \mathrm{FC}_{\lambda,g})
\}
|
}{
|\mathcal{G}^{\cap}_{\lambda}|
}.
\]

\item \textbf{Log Fold-change Spearman Correlation (LFCSpear).}

\[
\text{LFCSpear} =
\frac{1}{M}
\sum_{\lambda=1}^{M}
\mathrm{SpearmanR}
\left(
(\log \mathrm{FC}_{\lambda,g})_{g \in \mathcal{G}^{\mathrm{DE}}_{\lambda}},
(\widehat{\log \mathrm{FC}}_{\lambda,g})_{g \in \mathcal{G}^{\mathrm{DE}}_{\lambda}}
\right).
\]

\item \textbf{ROC-AUC (AUROC).}

Genes in $\mathcal{G}^{\mathrm{DE}}_{\lambda}$ are treated as positives and the remaining genes as negatives.
Predicted significance scores are defined as the negative log-transformed adjusted $p$-values.
AUROC is computed for each perturbation and averaged across perturbations.

\item \textbf{PR-AUC (AUPRC).}

This metric is calculated in the same way as AUROC but summarizes the precision--recall trade-off instead of the ROC curve.



\end{itemize}

\section{Supplementary ablation and optimization analysis}
\label{sec:ablation_results}

\noindent\textbf{Three design choices determine whether SCALE learns treatment-specific responses instead of simple averages.}
We tested the 184M-parameter model while changing one design choice at a time. The tests varied how condition inputs are combined, what the JiT model predicts and how its loss is calculated, and how the starting distribution is constructed.

\begin{figure}[t]
    \centering
    \includegraphics[width=0.9\textwidth]{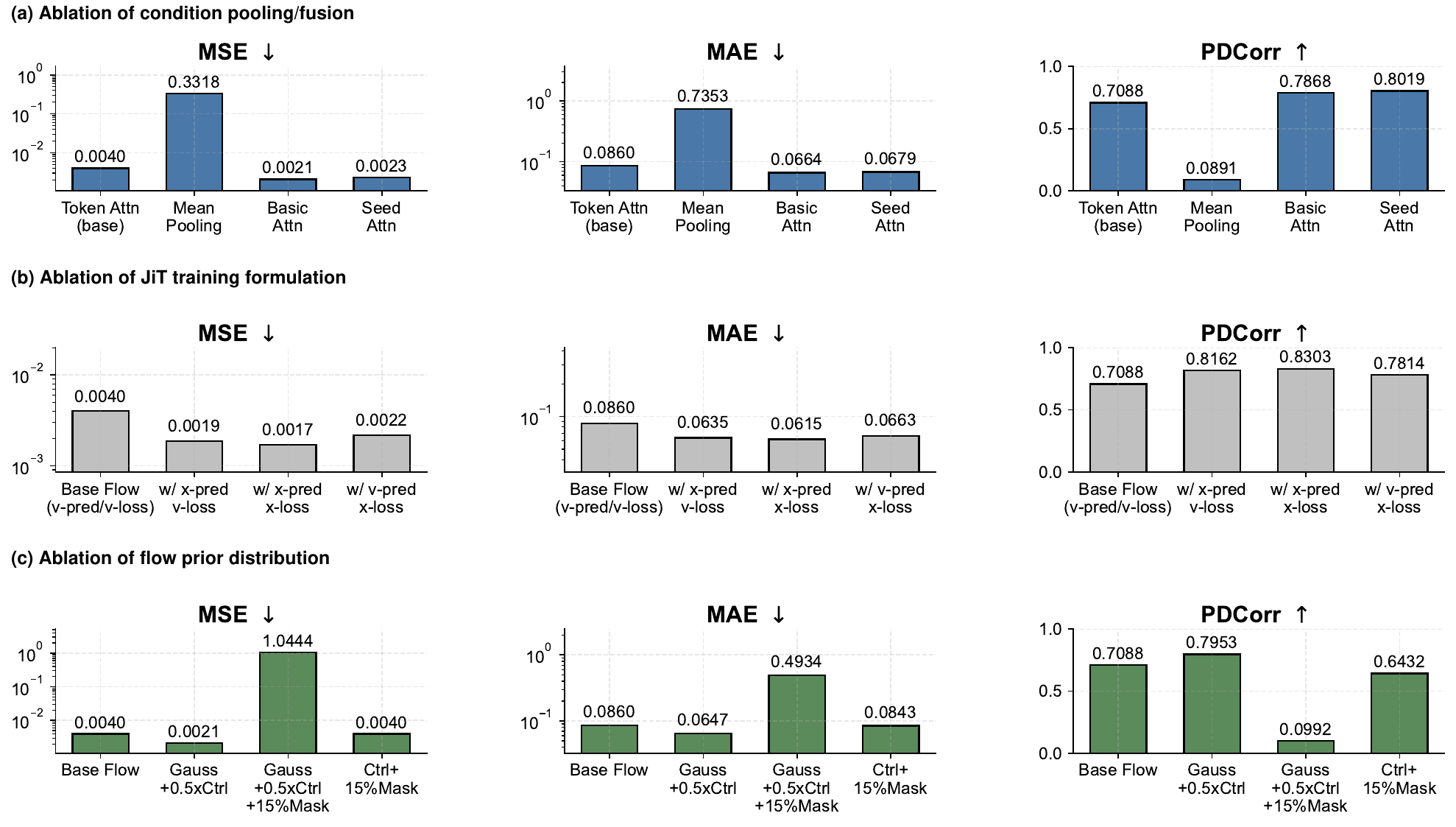}
    \caption{\textbf{Supplementary ablation study (a, condition aggregation; b, JiT formulation; c, start distribution).}
    We test three design choices in {\ProjectName}: how condition inputs are combined, how JiT is trained and how the starting distribution is constructed.
    (a) Mean pooling causes a clear performance drop, whereas basic and seed attention perform substantially better.
    (b) JiT improves over the base flow formulation, with direct endpoint prediction and endpoint loss (\textit{x-pred/x-loss}) giving the strongest overall performance.
    (c) Mixing Gaussian noise with the control distribution improves learning, whereas masking lowers performance. A harder starting point can help, but removing too much information in latent space is harmful.}
    \label{fig:ablation}
\end{figure}

\noindent\textbf{Learned attention preserves treatment information that is lost by simple averaging.}
Let $\{h_i\}_{i=1}^{N}$ denote token-level condition representations. We write the pooled condition embedding as
\begin{equation}
c = \sum_{i=1}^{N} \alpha_i h_i,
\qquad
\sum_{i=1}^{N}\alpha_i = 1,
\end{equation}
where each method uses a different weighting rule. Mean pooling gives every token the same weight $\alpha_i=1/N$. Token attention learns a separate weight for each token, and seed attention uses a learned query to combine the tokens.

Figure~\ref{fig:ablation}(a) shows that uniform mean pooling reduces PDCorr from 0.7088 to 0.0891, indicating that simple averaging removes treatment-specific information. Basic attention improves PDCorr to 0.7868, and seed attention raises it further to 0.8019. Learned attention is therefore important for combining different condition inputs.

\noindent\textbf{Directly predicting the treated endpoint gives the best JiT result.}
We varied both what the model predicts and how the loss is calculated. The model predicts either the treated endpoint $\hat{Z}_1$ (\textit{x-pred}) or the change from control $\hat{U}$ (\textit{v-pred}). It is trained against either the endpoint (\textit{x-loss}) or the change (\textit{v-loss}). The base flow formulation uses \textit{v-pred/v-loss}.

All JiT variants outperform the base formulation, but by different amounts (Figure~\ref{fig:ablation}b). The best result comes from \textit{x-pred/x-loss}, which raises PDCorr from 0.7088 to 0.8303 and lowers MSE and MAE. \textit{x-pred/v-loss} also performs well (0.8162), whereas \textit{v-pred/x-loss} is lower (0.7814) but remains above the baseline. Direct endpoint prediction is therefore more stable than predicting an intermediate velocity target in the current latent space.

Many responses in this benchmark are small, structured changes from control. Direct endpoint prediction may therefore match the evaluation target more closely than a loss on interpolated states. This result does not show that flow models are unnecessary in general. A shared vector field may still help with larger changes, more complex paths or explicit effects of time, dose and treatment combinations. It shows only that a multistep flow should not be assumed to beat direct endpoint prediction when the observed change is short.

\noindent\textbf{Mixing Gaussian noise with control cells improves learning, but masking removes too much information.}
We next replaced the control-based starting distribution with a mixture of Gaussian noise and control cells. When the start and endpoint are already close, the model may rely on a short interpolation instead of learning a response that transfers to new settings. A more different starting point makes this shortcut harder.

Compared with the base flow model, a Gaussian $+\,0.5\times$ control start raises PDCorr from 0.7088 to 0.7953 and lowers MSE and MAE (Figure~\ref{fig:ablation}c). However, masking 15\% of the Gaussian-control start causes performance to collapse, and masking the control start alone reduces PDCorr to 0.6432. A Gaussian-control mixture can help, but random masking removes too much information from the compressed latent representation.

\noindent\textbf{SCALE works best when condition attention, endpoint training and the starting distribution are considered together.}
These tests show that a low transport loss alone does not prove that the model learned a treatment response that transfers. The way conditions are combined, the prediction target and the starting distribution must be evaluated together against simple average-response and short-interpolation shortcuts.

\section{Supplementary scaling analysis}
\label{sec:scaling}

\noindent\textbf{Increasing SCALE from 184M to 280M parameters does not improve the measured biological responses.}
We evaluate both models with the complete h5ad-based Cell-Eval pipeline on the Tahoe validation and test splits.

\begin{table*}[ht]
\centering
\caption{
Scaling study of {\ProjectName} under
\emph{strict Cell-Eval} on Tahoe using the full h5ad-based evaluation pipeline.
}
\label{tab:scaling_study}
\resizebox{\textwidth}{!}{
\begin{tabular}{llcccccc}
\toprule
\textbf{Protocol} & \textbf{Eval Dataset} & \textbf{Params}  & \textbf{MSE}$\downarrow$ & \textbf{MAE}$\downarrow$ & \textbf{PDCorr}$\uparrow$ & \textbf{DEOver}$\uparrow$ & \textbf{LFCSpear}$\uparrow$ \\
\midrule
Strict Cell-Eval & Tahoe val  & 184M  & 0.0020 & 0.0058 & 0.9589 & 0.8026 & 0.8876 \\
Strict Cell-Eval & Tahoe val  & 280M  & 0.0020 & 0.0061 & 0.9556 & 0.7975 & 0.8742 \\
Strict Cell-Eval & Tahoe test & 184M  & 0.0020 & 0.0058 & 0.9526 & 0.8055 & 0.8758 \\
Strict Cell-Eval & Tahoe test & 280M  & 0.0020 & 0.0060 & 0.9502 & 0.8008 & 0.8669 \\
\bottomrule
\end{tabular}
}
\end{table*}

\noindent\textbf{The 184M model matches or slightly exceeds the 280M model on Tahoe validation and test data.}
We compared the selected 184M and 280M {\ProjectName} models using MSE, MAE, PDCorr and metrics for differentially expressed genes. On Tahoe \textit{val}, the 184M model slightly exceeds the 280M model in PDCorr (0.9589 versus 0.9556), DE Overlap (0.8026 versus 0.7975) and DE LFC Spearman (0.8876 versus 0.8742). The same pattern appears on Tahoe \textit{test}, where the 184M model reaches 0.9526 versus 0.9502 in PDCorr and 0.8055 versus 0.8008 in DE Overlap. The 280M model remains competitive, but it provides no reliable gain on these biological metrics under the current data, training procedure and evaluator. This result does not establish a general scaling law.

\end{document}